\documentclass{fairmeta}

\usepackage{microtype}
\usepackage{xcolor}
\usepackage{amsmath,amssymb,mathtools}
\usepackage{booktabs,multirow,tabularx}
\usepackage{array,float}
\usepackage{etoolbox}
\usepackage{colortbl}
\usepackage{makecell}
\usepackage{graphicx}
\usepackage{caption,subcaption}
\usepackage{enumitem}
\usepackage{titlesec}
\usepackage{natbib}
\usepackage{url}
\usepackage{hyperref}
\usepackage{bookmark}

\colorlet{ink}{metafg}
\colorlet{accent}{metablue}
\definecolor{accentlight}{HTML}{EAF4F6}
\definecolor{oursrow}{HTML}{EEF5FC}
\definecolor{zebrarow}{HTML}{F6F8FA}
\definecolor{rulegray}{HTML}{B8C5CC}

\setlist[itemize]{leftmargin=1.35em,itemsep=0.22em,topsep=0.25em}
\setlist[enumerate]{leftmargin=1.5em,itemsep=0.22em,topsep=0.25em}
\AtBeginEnvironment{table}{\footnotesize}
\AtBeginEnvironment{table*}{\footnotesize}
\ExplSyntaxOn
\box_new:N \l_paired_table_box
\dim_new:N \l_paired_first_dim
\dim_new:N \l_paired_second_dim
\NewDocumentCommand{\fitpairedtable}{m m +m}{
  \group_begin:
  \hbox_set:Nn \l_paired_table_box
    {\renewcommand{\arraystretch}{1}\resizebox{#1}{!}{#3}}
  \dim_set:Nn \l_paired_first_dim
    {\box_ht:N \l_paired_table_box + \box_dp:N \l_paired_table_box}
  \hbox_set:Nn \l_paired_table_box
    {\renewcommand{\arraystretch}{2}\resizebox{#1}{!}{#3}}
  \dim_set:Nn \l_paired_second_dim
    {\box_ht:N \l_paired_table_box + \box_dp:N \l_paired_table_box}
  \cs_set:Npx \arraystretch {\fp_eval:n {
    1 + (\dim_to_fp:n {#2} - \dim_to_fp:n {\l_paired_first_dim}) /
    (\dim_to_fp:n {\l_paired_second_dim} - \dim_to_fp:n {\l_paired_first_dim})}}
  \hbox_set:Nn \l_paired_table_box {\resizebox{#1}{!}{#3}}
  \cs_set:Npx \arraystretch {\fp_eval:n {\arraystretch +
    (\dim_to_fp:n {#2} - \dim_to_fp:n {
      \box_ht:N \l_paired_table_box + \box_dp:N \l_paired_table_box}) /
    (\dim_to_fp:n {\l_paired_second_dim} - \dim_to_fp:n {\l_paired_first_dim})}}
  \resizebox{#1}{!}{#3}
  \group_end:
}
\ExplSyntaxOff

\newcommand{\method}{\alias}

\newcommand{\alias}{$\mathcal{F}^{2}$}

\hypersetup{
  pdftitle={Foresight Forcing: Learning Spatial Foresight for Robotic Manipulation},
  pdfauthor={Mishuo Han, Zhixuan Liang, Zhaoyuan Wan, Wendi Chen, Jiarui Chen, Xinkai Wang, Yue Ma, Yu Zhou, Junting Chen, Yao Mu, Jun Lv, Chuan Wen, Cewu Lu},
  colorlinks=true,
  linkcolor=ink,
  citecolor=accent,
  urlcolor=accent
}

\title{Foresight Forcing: Learning Spatial Foresight for Robotic Manipulation}
\newcommand{\affmark}[1]{\raisebox{0.5ex}{\footnotesize #1}}
\newcommand{\paperauthor}[3]{\mbox{\href{https://openreview.net/profile?id=\string~#2}{\textbf{#1}}\affmark{#3}}}
\newcommand{\paperauthorblock}{%
  \small
  \paperauthor{Mishuo Han}{Mishuo_Han1}{1,3,4,*},
  \paperauthor{Zhixuan Liang}{Zhixuan_Liang2}{2,*,\ensuremath{\dagger}},
  \paperauthor{Zhaoyuan Wan}{Zhaoyuan_Wan2}{1,4,*},
  \paperauthor{Wendi Chen}{Wendi_Chen1}{1,3,4,*},\linebreak
  \paperauthor{Jiarui Chen}{Jiarui_Chen4}{3,4,5},
  \paperauthor{Xinkai Wang}{Xinkai_Wang1}{3,4},
  \paperauthor{Yue Ma}{Yue_Ma2}{6},
  \paperauthor{Yu Zhou}{Yu_Zhou21}{1,3,4},
  \paperauthor{Junting Chen}{Junting_Chen2}{7},\linebreak
  \paperauthor{Yao Mu}{Yao_Mu1}{1},
  \paperauthor{Jun Lv}{Jun_Lv2}{4},
  \paperauthor{Chuan Wen}{Chuan_Wen1}{1},
  \paperauthor{Cewu Lu}{Cewu_Lu3}{1,3,4,\ensuremath{\dagger}}\par
  \vspace{0.6em}
  {\footnotesize
  \affmark{1}\,\href{https://sjtu.edu.cn}{Shanghai Jiao Tong University}\quad
  \affmark{2}\,\href{https://www.hku.hk}{The University of Hong Kong}\par
  \affmark{3}\,\href{https://sii.edu.cn}{Shanghai Innovation Institute}\quad
  \affmark{4}\,\href{https://www.noematrix.ai}{Noematrix}\quad
  \affmark{5}\,\href{https://www.fudan.edu.cn}{Fudan University}\par
  \affmark{6}\,\href{https://ust.hk}{The Hong Kong University of Science and Technology}\quad
  \affmark{7}\,\href{https://u.nus.edu}{National University of Singapore}\par
  \vspace{0.35em}
  \affmark{*}\,Equal contribution.\quad
  \affmark{\ensuremath{\dagger}}\,Corresponding authors.\par}%
}
\renewcommand{\authorlist}{\paperauthorblock}
\abstract{%
Robotic manipulation policies can improve action prediction by anticipating how objects move and how their spatial relationships change. World Action Models (WAMs) learn to predict future visual observations, but their training objectives typically leave the underlying 3D geometric transitions implicit.
To alleviate this problem, we introduce Foresight Forcing (\method{}), a WAM that learns a spatial transition representation from changes in a pretrained 3D feature space.
During training, a frozen 3D foundation model extracts patch-wise feature differences from aligned current and future observations that provide dense supervision for geometry-aware feature transitions. We distill these targets into spatial transition queries, which are progressively refined from the current observation, language instruction, and proprioceptive state, and use them alongside current visual features to generate actions. At deployment, the policy predicts the queries without future observations or the 3D teacher.
Experiments on LIBERO, LIBERO-Plus, and RoboTwin~2.0 demonstrate strong manipulation performance and improved robustness, including 95.01\% average success on RoboTwin~2.0 Full. On six real-world manipulation tasks, \method{} achieves a mean task score of 85.8\%. 
\par\vspace{0.05em}\noindent\textbf{Website:}~\url{https://anson2333.github.io/foresight-forcing}
}

\begin{document}
\maketitle

\begin{figure}[H]
    \centering
    \includegraphics[width=\textwidth]{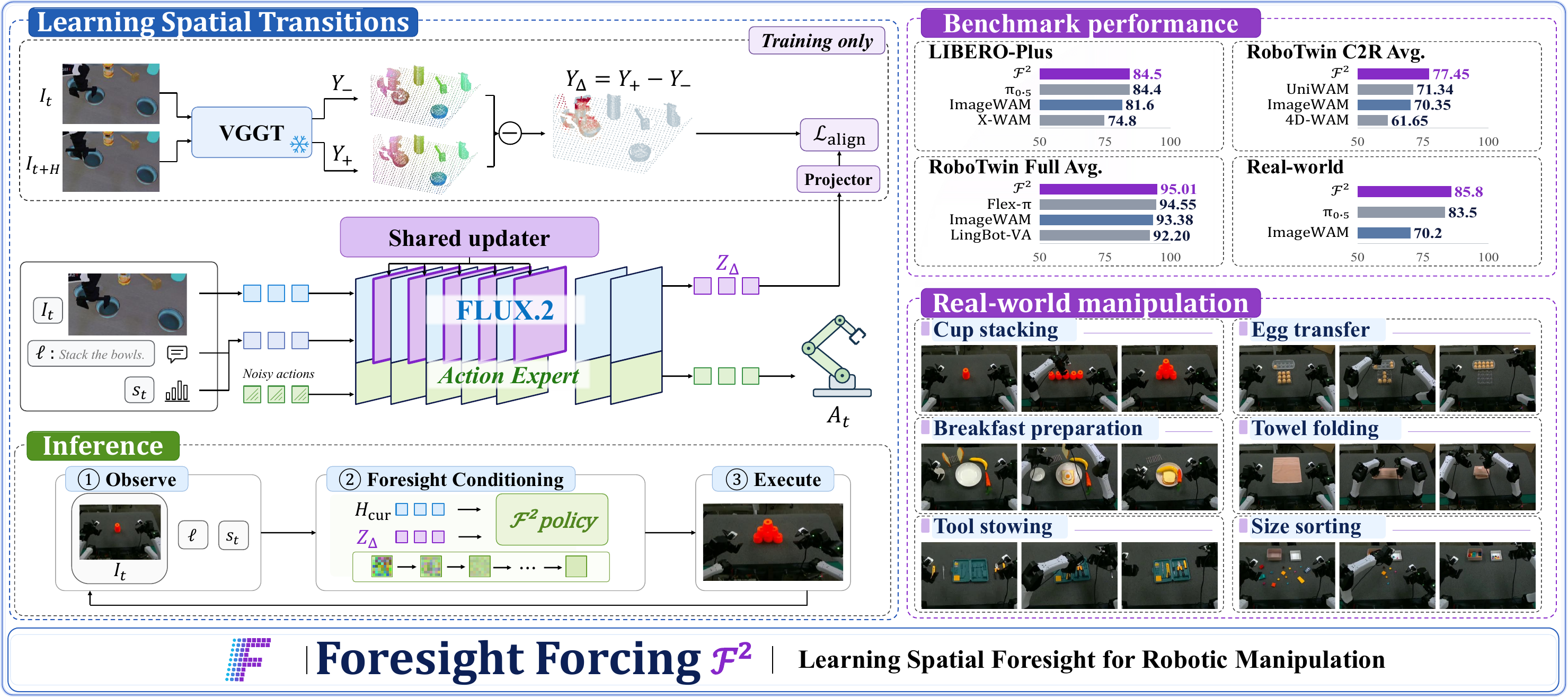}
    \captionsetup{font=footnotesize,skip=4pt}
    \caption{\textbf{Foresight Forcing (\alias) at a glance.} \alias{} learns spatial transition queries from correspondence-aligned changes in 3D teacher features. During training, a frozen VGGT extracts features from paired observations to supervise the queries. At inference, the queries are predicted from the current observation, instruction, and proprioceptive state and, together with current visual features, condition action generation without the teacher or future observations.}
    \label{fig:teaser}
\end{figure}

\clearpage
\section{Introduction}
\label{sec:intro}

Robotic manipulation has progressed from task-specific systems designed for controlled environments towards generalist policies that operate across diverse objects, instructions, and settings. A general-purpose robot must interpret task instructions, perceive objects in its surroundings, and execute precise actions under varying conditions. Recent advances in large-scale robot learning and pretrained vision-language models have enabled Vision-Language-Action (VLA) policies to map visual observations and natural-language instructions to robot actions~\cite{zitkovich2023rt2,kim2025openvla,24pi0,intelligence2025pi05,kim2025openvlaoft}.
However, action labels do not directly supervise the visual outcomes of those actions. To incorporate this information into policy learning, World Action Models (WAMs) build on visual generative models and couple action learning with world modeling objectives~\cite{imagewam,cen2025worldvla,bi2026motus,yuan2026fastwam}. The resulting visual prediction task provides an additional training signal alongside action supervision and enables robot policies to benefit from pretrained generative representations.

Most WAMs learn how a scene evolves by predicting future RGB observations or visual latents. These objectives encode task-relevant spatial relationships alongside object appearance, background, and viewpoint, leaving the geometric transitions that matter for control implicit. 
Geometry Forcing shows that pretrained 3D features can improve the geometric consistency of video generation~\cite{wu2025geometryforcing}, and geometry-aware WAMs have explored geometric supervision for manipulation~\cite{ma2026geosemwam,zhang2026mecowam,li2026wam4d}.
For robot control, the 3D changes needed to advance the task provide a natural target for this supervision.
Placing an object requires changing its pose relative to a target, while opening a drawer requires changing the configuration of an articulated object. 
Modeling such transitions gives the policy a way to anticipate how the scene needs to change when generating actions.

Therefore, we introduce Foresight Forcing (\method{}), a WAM that fuses correspondence-aligned transitions in 3D geometry feature space of foundation models into spatial transition queries for action generation, rather than directly predicting explicit 3D motions (Figure~\ref{fig:teaser}). During training, a frozen VGGT teacher~\cite{wang2025vggt} extracts patch features from an aligned observation pair, and their temporal differences serve as dense targets for the queries. A shared updater progressively refines the queries after each double block of the visual backbone using information from the current observation, language instruction, and proprioceptive state. We supervise the queries to match the direction and scale of the teacher targets and feed the refined queries directly to the action expert. At inference, the policy predicts the queries from its current inputs without access to future observations or the teacher. We evaluate \method{} on LIBERO, LIBERO-Plus, RoboTwin~2.0, and real-world manipulation tasks to assess manipulation performance, robustness, and inference efficiency. \method{} achieves state-of-the-art performance across simulation and real robot.

\begingroup
\setlength{\parskip}{0pt}
Our contributions are summarized as follows:
\begin{itemize}[topsep=0pt,partopsep=0pt,itemsep=0pt,parsep=0pt]
    \item \textbf{Transition-centric geometric supervision.} We introduce correspondence-aligned 3D teacher feature differences as supervision for spatial transition learning for action generation in a World Action Model. 
    \item \textbf{Teacher-guided spatial transition learning for action generation.} We introduce dedicated latent queries that a shared updater progressively refines across the visual backbone. During training, we align the refined queries with the frozen 3D teacher targets in direction and scale. The queries directly condition the action expert.
    \item \textbf{Manipulation performance and robustness.} Evaluations on LIBERO, LIBERO-Plus, and RoboTwin~2.0 demonstrate strong manipulation performance and improved robustness, with state-of-the-art average success rates on RoboTwin~2.0 Full and Clean-to-Random. On six real-world tasks, \method{} achieves a mean task score of 85.8\%, exceeding ImageWAM and $\pi_{0.5}$ by 15.6 and 2.3 percentage points, respectively.
\end{itemize}
\endgroup

\begin{figure*}[t]
    \centering
    \includegraphics[width=\textwidth]{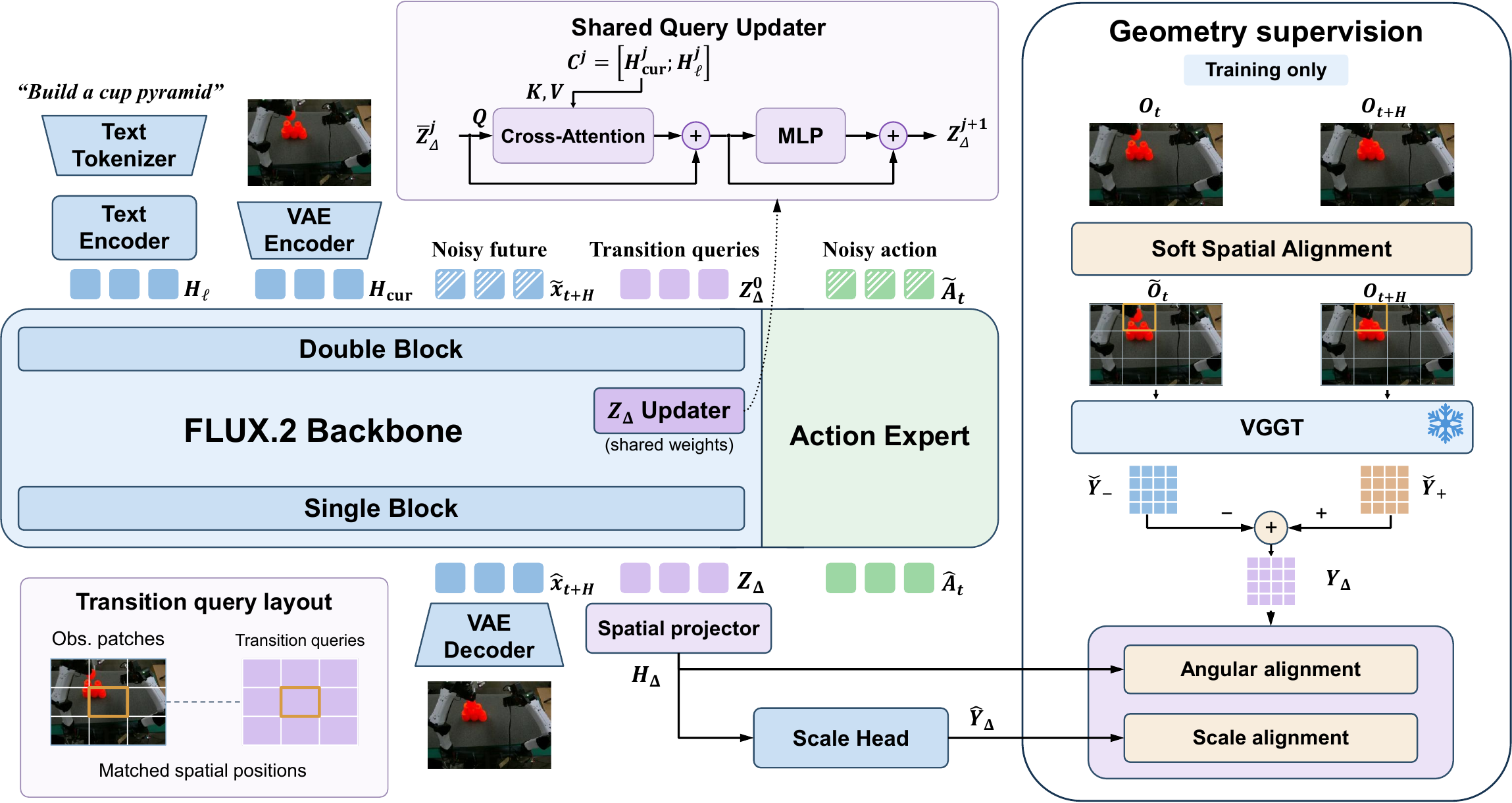}
    \caption{\textbf{\alias{} architecture.} A frozen VGGT provides feature transition targets on the student grid for spatial queries that condition the Action Expert. Insets show the shared updater and spatial query layout. VGGT and future observations are used only during training.}
    \label{fig:overview}
\vspace{-10pt}
\end{figure*}

\section{Related Work}
\label{sec:related}

\paragraph{World Action Models.}
World Action Models (WAMs) jointly learn action generation and future visual prediction through autoregressive video-action modeling, shared latent representations, compact future conditions, or image-based prediction~\cite{cen2025worldvla,bi2026motus,lingbotva,su2026worldguidance,yuan2026fastwam,imagewam}. Visual prediction provides supervision for control, even when future images are not generated during execution. Building on ImageWAM, we study a complementary signal: the spatial transition between current and future observations.

\paragraph{3D Representations for Visual Generation.}
Feed-forward 3D models recover spatial structure and dense features from images. DUSt3R and MASt3R establish pairwise geometry and correspondences~\cite{wang2024dust3r,leroy2024mast3r}; subsequent models extend inference to multiple views and image streams~\cite{yang2025fast3r,wang2025cut3r,wang2025vggt,wang2025pi3,lin2025da3,wang2026vggtomega}. REPA aligns diffusion representations with pretrained visual features~\cite{yu2025repa}, while geometric features guide consistent video generation and image editing~\cite{wu2025geometryforcing,liu2026camgeo,ma2026groupediting}. We use a frozen VGGT to encode paired observations and supervise the policy with correspondence-aligned differences between its current and future features.

\paragraph{Geometry in Action Learning.}
Existing action models incorporate geometry either as an observation or predicted spatial representation, or as supervision for learned features. The former includes geometric inputs, point flow, trajectories, future pointmaps, and spatially structured actions~\cite{huang2026pointworld,park2026pointwam,wang2026unitas,chen2026mvgwam,yan2026flexpi,guo2026xwam,zhang2026mecowam}. The latter uses geometric objectives or distilled priors to shape policy representations~\cite{li2026spatial,ma2026geosemwam,li2026wam4d}. Foresight Forcing instead supervises dedicated spatial queries with correspondence-aligned temporal differences in 3D teacher features. The queries are predicted from current inputs and directly condition the action expert.

\section{Method}
\label{sec:method}

World Action Models (WAMs) jointly predict future observations and robot actions. Building on this framework, \alias{} introduces spatial transition queries that encode task-relevant geometric changes over the action horizon. During training, a frozen 3D teacher supplies feature differences between aligned current and future observations as dense targets. The queries condition action generation alongside current visual features; neither future observations nor the teacher is required at deployment (Figure~\ref{fig:overview}). We formulate the task in Sec.~\ref{sec:formulation}, construct the geometric transition target in Sec.~\ref{sec:teacher}, explain how the queries learn from it in Sec.~\ref{sec:queries}, and detail training and deployment in Sec.~\ref{sec:conditioning}.

\subsection{Problem Formulation}
\label{sec:formulation}

We consider language-conditioned robotic manipulation from multi-view observations. At time $t$, the policy receives images $O_t=\{I_t^v\}_{v=1}^{V}$ from $V$ cameras, a language instruction $\ell$, and the robot's proprioceptive state $s_t$. It predicts an action chunk $A_t=(a_t,\ldots,a_{t+H-1})$ of horizon $H$. During training, we pair $O_t$ with the future observation $O_{t+H}$, which the policy cannot access at deployment, to construct transition supervision. We use $H=16$ in our experiments.

Rather than predicting explicit 3D motion, \alias{} augments a WAM with spatial transition queries that represent task-relevant geometric changes over the action horizon:
\begin{equation}
    Z_\Delta=F_\theta(O_t,\ell,s_t).
    \label{eq:transition-prediction}
\end{equation}
The model is jointly trained for future-observation modeling and transition-conditioned action generation:
\begin{equation}
\begin{aligned}
    O_{t+H} &\sim p_\theta(\,\cdot\mid O_t,\ell,s_t),\\
    A_t &\sim \pi_\theta(\,\cdot\mid O_t,\ell,s_t,Z_\Delta).
\end{aligned}
\label{eq:policy}
\end{equation}
Here $F_\theta$, $p_\theta$, and $\pi_\theta$ denote the transition predictor, future-observation model, and action policy, respectively.

\paragraph{Notation.}
In Figure~\ref{fig:overview}, $x_{t+H}$ is the future-image VAE latent; tildes and hats on $x,A$ denote noisy inputs and predictions, respectively. We use $\widetilde O_t=\{\widetilde I_t^v\}_{v=1}^{V}$ for aligned current images, $(H_\Delta)_p=q_p$, and $(\widehat Y_\Delta)_p=S_\theta(\widehat q_p)$, with $\widehat q_p$ denoting normalized $q_p$ (Sec.~\ref{sec:queries}).

\subsection{Constructing Transition Supervision}
\label{sec:teacher}

\paragraph{Correspondence-aligned 3D features.}
We use a frozen VGGT~\cite{wang2025vggt} as the 3D teacher. For each view, we align the current image to the future-image grid, obtaining $\widetilde I_t^v$ (Appendix~\ref{sec:implementation}). VGGT jointly encodes the aligned current image and the future image to produce patch features:
\begin{equation}
    (Y_-^v,Y_+^v)=G_{\mathrm{patch}}
    \!\left(\widetilde I_t^v,I_{t+H}^v\right).
    \label{eq:vggt-features}
\end{equation}
We resize the resulting feature grids to the token grid of our FLUX.2 [klein] base 4B backbone~\cite{flux-2-2025}, obtaining $\check Y_-^v,\check Y_+^v\in\mathbb{R}^{N_v\times D}$, where $N_v$ is the number of visual tokens in view $v$. These features provide the spatially aligned representations used to construct the geometric transition target.

\paragraph{Geometric feature transition.}
The aligned feature pair describes the current and future observations on a shared spatial grid. We represent their transition through the patch-wise feature difference
\begin{equation}
    Y_{\Delta}^v
    =\check Y_{+}^v-\check Y_{-}^v
    \in\mathbb{R}^{N_v\times D}.
    \label{eq:teacher-delta}
\end{equation}
The subtraction is performed at corresponding locations in the feature grids. The resulting target captures changes in 3D teacher feature space after image correspondence alignment and provides dense supervision for the spatial transition queries. For simplicity, we omit the view index $v$ in the following.

\subsection{Learning Spatial Transitions}
\label{sec:queries}

\paragraph{Spatial transition queries.}
The future observation and teacher target are available only during training. To predict anticipated transitions at deployment, we introduce learnable queries $Z_\Delta^0\in\mathbb{R}^{N\times d}$ aligned to the $N=\sum_{v=1}^{V} N_v$ visual token locations across all $ V $ camera views (e.g., $N=392$ on LIBERO's two-view $14\times 28$ grid), where $d$ is the query width. The query grid follows the spatial layout of the visual tokens, allowing each query to predict the feature transition at a corresponding image location.

Within the joint Mixture-of-Transformers (MoT) backbone of FLUX.2 and the Action Expert (Figure~\ref{fig:overview}), $Z_\Delta$ is processed alongside the conditioning ($H_\ell$), current-image ($H_{\mathrm{cur}}$), future-image (training only), and action tokens. Inside every double and single block, native attention for $Z_\Delta$ is restricted to self-attention alone---masking out future, current, instruction, and action tokens so no future-frame information can leak during training. After double block $j$ outputs post-block states $\bar Z_\Delta^j$, a shared updater $U_\theta$ (Figure~\ref{fig:overview}, top inset) reads only current visual states $H_{\mathrm{cur}}^j$ and conditioning states $H_\ell^j$:
\begin{equation}
    Z_\Delta^{j+1}=U_\theta\!\left(\bar Z_\Delta^j,
    [H_{\mathrm{cur}}^j;H_\ell^j]\right).
    \label{eq:update}
\end{equation}
$H_\ell^j$ packs the language instruction and encoded proprioceptive state tokens (likewise masked from future tokens). $U_\theta$ applies pre-normalized residual cross-attention and a feed-forward network with parameters shared across double blocks. Subsequent single blocks continue refining $Z_\Delta$ via self-attention before teacher supervision, while the Action Expert directly attends to $Z_\Delta$ across the MoT blocks to condition action generation.

\paragraph{Geometric transition alignment.}
We project the transition queries into the teacher feature space and align them with the geometric transition targets. Let $q_p=P_\theta(Z_\Delta)_p\in\mathbb{R}^{D}$ and $y_{\Delta,p}$ denote the projected query and teacher target at location $p$, where $p$ indexes locations across all views. Following the direction-and-scale alignment objective of Geometry Forcing~\cite{wu2025geometryforcing}, we define
\begin{align}
    \mathcal L_{\mathrm{ang}}
    &=-\frac{1}{W}\sum_p w_p\,
      \cos_\epsilon\!\left(q_p,y_{\Delta,p}\right),
      \label{eq:angular}\\
    \mathcal L_{\mathrm{scale}}
    &=\frac{1}{WD}\sum_p w_p\,
      \left\|S_\theta(\widehat q_p)-y_{\Delta,p}\right\|_2^2,
      \label{eq:scale}
\end{align}
$S_\theta$ predicts teacher-space features from $\widehat q_p$, and the patch weights satisfy $W=\sum_p w_p$. The angular and scale objectives align feature directions and reconstruct teacher-space features, respectively. The patch weights account for alignment validity and transition magnitude; their definition, the projector architecture, and a feature-norm regularizer $\mathcal L_{\mathrm{rad}}$ are provided in Appendix~\ref{sec:implementation}.

\subsection{Training Strategy and Deployment}
\label{sec:conditioning}

\paragraph{Reference-condition dropout.}
Within the MoT attention mask, the Action Expert attends to the instruction, proprioceptive state, action tokens, and $Z_\Delta$ across blocks while staying masked from future tokens. Attention to $H_{\mathrm{cur}}$ is enabled with probability $q=0.5$ during training, exposing the Action Expert to both reference-assisted and transition-only conditioning. At inference, future tokens, VGGT, and distillation heads are omitted, and both $H_{\mathrm{cur}}$ and $Z_\Delta$ remain visible. Complete attention masks and token-visibility rules are provided in Appendix~\ref{sec:implementation}.

\paragraph{Optimization.}
We jointly optimize future-image prediction, action generation, and geometric transition alignment:
\begin{align}
    \mathcal L
    ={}&\lambda_{\mathrm{img}}\mathcal L_{\mathrm{img}}
       +\lambda_{\mathrm{act}}\mathcal L_{\mathrm{act}}
       +\lambda_{\mathrm{geo}}\rho(k)\mathcal L_{\mathrm{geo}},\\
    \mathcal L_{\mathrm{geo}}
    ={}&\lambda_{\mathrm{ang}}\mathcal L_{\mathrm{ang}}
       +\lambda_{\mathrm{scale}}\mathcal L_{\mathrm{scale}}
       +\lambda_{\mathrm{rad}}\mathcal L_{\mathrm{rad}}.
    \label{eq:total}
\end{align}
Here $\mathcal L_{\mathrm{img}}$ and $\mathcal L_{\mathrm{act}}$ are the image and action flow-matching losses~\cite{lipman2023flow}, following ImageWAM~\cite{imagewam}, and $\rho(k)$ gradually increases the weight of geometric supervision during training. Details are provided in Appendix~\ref{sec:implementation}.

\section{Experiments}
\label{sec:experiments}
We evaluate manipulation performance, robustness to distribution shifts, the contributions of individual components, real-world applicability, and inference cost.

\subsection{Experimental Setup}
\paragraph{Benchmarks.}
We evaluate \method{} on LIBERO~\cite{LIBERO23}, LIBERO-Plus~\cite{fei25libero-plus}, RoboTwin~2.0~\cite{chen2025robotwin}, and real-world manipulation tasks. LIBERO is a single-arm simulation benchmark; we evaluate its Spatial, Object, Goal, and Long suites, each containing 10 tasks. LIBERO-Plus tests robustness across seven perturbation dimensions: object layout, camera viewpoint, robot initial state, language instruction, lighting, background, and observation noise. RoboTwin~2.0 comprises 50 dual-arm manipulation tasks in simulated tabletop scenes, including object placement, stacking, tool use, and articulated-object interaction. We evaluate both its Full and Clean-to-Random settings under clean and randomized scenes. For real-world evaluation, we compare three methods on six manipulation tasks.

\paragraph{Models and evaluation.}
Following ImageWAM~\cite{imagewam}, we train \method{} with geometric transition supervision added to ImageWAM after fine-tuning on the corresponding benchmark demonstrations. For LIBERO, LIBERO-Plus, and RoboTwin Full, we do not use robot pretraining. For RoboTwin Clean-to-Random, we use robot pretraining on InternData-A1~\cite{tian2025interndata} before fine-tuning on clean demonstrations. As a control, we also continue fine-tuning ImageWAM without geometric supervision under the same settings; evaluation shows no consistent improvement, with slight decreases in some settings. We compare \method{} with ImageWAM and representative VLA and WAM methods using task success rates. Training configurations are provided in Appendix~\ref{sec:implementation}.

\subsection{In-Distribution Performance}
\label{sec:comparison}
We first examine whether \method{} maintains strong standard manipulation performance when the evaluation follows the training distribution.

\paragraph{LIBERO.}

Table~\ref{tab:comparison-libero} reports success on the four standard suites. Each suite contains 10 tasks, and we evaluate each task over 50 trials. \method{} achieves an average success rate of 98.5\%, compared with 98.4\% for ImageWAM and matching the leading methods at the reported precision. As these methods approach the success-rate ceiling, standard LIBERO offers limited headroom for distinguishing their performance.

\paragraph{RoboTwin 2.0.}

In the RoboTwin 2.0 Full setting, \method{} achieves the highest clean, randomized, and average success rates in Table~\ref{tab:comparison-robotwin}, without robot-policy pretraining. It reaches 94.90\% and 95.12\% in the clean and randomized settings, respectively, improving over ImageWAM by 1.70 and 1.56 percentage points. Its average success is 95.01\%,
\begin{table}[t]
\centering
\begin{minipage}[t]{0.485\textwidth}
\centering
\begin{minipage}[t][40pt][t]{\linewidth}
\captionof{table}{\textbf{LIBERO benchmark.} Success rates (\%) on the four standard suites. Geo.\ denotes geometric inputs or supervision.}
\label{tab:comparison-libero}
\end{minipage}\par\vspace{2pt}
\footnotesize
\setlength{\tabcolsep}{2.2pt}
\renewcommand{\arraystretch}{1.205}
\begin{tabularx}{\linewidth}{>{\raggedright\arraybackslash}Xcccccc}
\toprule
Method & Geo. & Spat. & Obj. & Goal & Long & Avg. \\
\midrule
\multicolumn{7}{l}{\textit{Vision-Language-Action Models}} \\
OpenVLA-OFT~\cite{kim2025openvlaoft} & $\times$ & 97.6 & 98.4 & 97.9 & 94.5 & 97.1 \\
$\pi_{0.5}$~\cite{intelligence2025pi05} & $\times$ & 98.8 & 98.2 & 98.0 & 92.4 & 96.9 \\
Spatial Forcing~\cite{li2026spatial} & $\checkmark$ & \underline{99.4} & 99.6 & \textbf{98.8} & 96.0 & \textbf{98.5} \\
\midrule
\multicolumn{7}{l}{\textit{World Action Models}} \\
LingBot-VA~\cite{lingbotva} & $\times$ & 98.5 & 99.6 & 97.2 & \textbf{98.5} & \textbf{98.5} \\
Motus~\cite{bi2026motus} & $\times$ & 96.8 & \underline{99.8} & 96.6 & 97.6 & 97.7 \\
Fast-WAM~\cite{yuan2026fastwam} & $\times$ & 98.2 & \textbf{100.0} & 97.0 & 95.2 & 97.6 \\
MECo-WAM~\cite{zhang2026mecowam} & $\checkmark$ & 98.8 & \textbf{100.0} & 98.2 & 95.8 & 98.2 \\
Flex-$\pi$ (full joint)~\cite{yan2026flexpi} & $\checkmark$ & \textbf{99.6} & \underline{99.8} & \underline{98.6} & 96.0 & \textbf{98.5} \\
ImageWAM~\cite{imagewam} & $\times$ & 97.2 & 99.2 & \textbf{98.8} & \underline{98.4} & \underline{98.4} \\
\rowcolor{oursrow}
\textbf{\method{}} & $\checkmark$ & \underline{99.4} & \textbf{100.0} & 97.8 & 96.6 & \textbf{98.5} \\
\bottomrule
\end{tabularx}
\end{minipage}\hfill
\begin{minipage}[t]{0.485\textwidth}
\centering
\begin{minipage}[t][40pt][t]{\linewidth}
\captionof{table}{\textbf{RoboTwin 2.0 Full.} Success rates (\%) on clean and randomized scenes. P.T.: policy pretraining; Geo.: geometric supervision/inputs.}
\label{tab:comparison-robotwin}
\end{minipage}\par\vspace{2pt}
\footnotesize
\setlength{\tabcolsep}{3.5pt}
\renewcommand{\arraystretch}{1.12}
\begin{tabularx}{\linewidth}{Xccccc}
\toprule
Method & P.T. & Geo. & Clean & Rand. & Avg. \\
\midrule
\multicolumn{6}{l}{\textit{Vision-Language-Action Models}} \\
$\pi_{0.5}$~\cite{intelligence2025pi05} & $\checkmark$ & $\times$ & 82.74 & 76.76 & 79.75 \\
\midrule
\multicolumn{6}{l}{\textit{World Action Models}} \\
Motus~\cite{bi2026motus} & $\checkmark$ & $\times$ & 88.66 & 87.02 & 87.84 \\
LingBot-VA~\cite{lingbotva} & $\checkmark$ & $\times$ & 92.90 & 91.50 & 92.20 \\
X-WAM~\cite{guo2026xwam} & $\checkmark$ & $\checkmark$ & 89.80 & 90.70 & 90.25 \\
WAM4D~\cite{li2026wam4d} & $\checkmark$ & $\checkmark$ & 93.80 & 89.90 & 91.80 \\
Flex-$\pi$ (full joint)~\cite{yan2026flexpi} & $\checkmark$ & $\checkmark$ & \underline{94.30} & \underline{94.80} & \underline{94.55} \\
Fast-WAM~\cite{yuan2026fastwam} & $\checkmark$ & $\times$ & 91.88 & 91.78 & 91.83 \\
GeoSem-WAM~\cite{ma2026geosemwam} & $\times$ & $\checkmark$ & 92.94 & 92.14 & 92.54 \\
MECo-WAM~\cite{zhang2026mecowam} & $\times$ & $\checkmark$ & 93.26 & 91.98 & 92.62 \\
ImageWAM~\cite{imagewam} & $\times$ & $\times$ & 93.20 & 93.56 & 93.38 \\
\rowcolor{oursrow}
\textbf{\method{}} & $\times$ & $\checkmark$ & \textbf{94.90} & \textbf{95.12} & \textbf{95.01} \\
\bottomrule
\end{tabularx}
\end{minipage}
\end{table}
compared with 94.55\% for Flex-$\pi$. The gains over ImageWAM in both settings suggest that spatial transition supervision benefits manipulation across the evaluated scene conditions. Per-task results are reported in Appendix~\ref{sec:robotwin-per-task}.

\subsection{Out-of-Distribution Robustness}
\label{sec:ablations-robustness}
We evaluate robustness to distribution shifts on LIBERO-Plus and RoboTwin Clean-to-Random.

\paragraph{LIBERO-Plus.}
For this evaluation, we use only the original LIBERO training demonstrations and do not incorporate the augmented LIBERO-Plus training data. As shown in Table~\ref{tab:comparison-libero-plus}, \method{} achieves an overall success rate of 84.5\%, compared with 81.6\% for ImageWAM. The largest improvements over ImageWAM occur under robot-state and camera perturbations, where \method{} improves by 8.7 and 5.4 percentage points, respectively. These results suggest that spatial transition supervision can help the policy handle changes in robot state and camera viewpoint. Meanwhile, $\pi_{0.5}$ and OpenWAM-$\alpha$ achieve stronger results under robot-state perturbations, indicating that these perturbations remain challenging.

\paragraph{RoboTwin Clean-to-Random.}
In the RoboTwin Clean-to-Random (C2R) setting, we train on clean demonstrations and evaluate on both clean and randomized scenes. As shown in Table~\ref{tab:comparison-robotwin-c2r}, \method{} achieves the highest average success rate among the compared methods, reaching 87.84\% on clean scenes and 67.06\% on randomized scenes, with an average of 77.45\%. Its improvement over ImageWAM in both conditions suggests that the learned transition representation remains useful when the evaluation scenes change. The gap between clean and randomized performance nevertheless shows that generalization from clean demonstrations remains challenging. Per-task results are reported in Appendix~\ref{sec:robotwin-per-task}. The 4D-WAM values in Table~\ref{tab:comparison-robotwin-c2r} follow the official RoboTwin 2.0 leaderboard.

\begin{table}[t]
\centering
\caption{\textbf{LIBERO-Plus robustness.} Success rates (\%) across seven perturbation dimensions without augmented training data. P.T.\ denotes embodied-policy pretraining; Cam.\ and Rob.\ denote camera and robot-state perturbations.}
\label{tab:comparison-libero-plus}
\footnotesize
\setlength{\tabcolsep}{6.5pt}
\renewcommand{\arraystretch}{1.14}
\begin{tabularx}{\textwidth}{Xcrrrrrrrr}
\toprule
Method & P.T. & Cam. & Rob. & Lang. & Light & Bg. & Noise & Layout & Overall \\
\midrule
\multicolumn{10}{l}{\textit{Vision-Language-Action Models}} \\
$\pi_{0.5}$~\cite{intelligence2025pi05} & $\checkmark$ & \underline{78.4} & \underline{73.6} & 80.8 & 96.2 & \underline{94.1} & 89.0 & \textbf{84.5} & \underline{84.4} \\
starVLA~\cite{starvla2026} & $\times$ & 64.0 & 57.2 & 85.8 & 93.8 & \textbf{96.0} & 81.0 & 77.3 & 77.8 \\
\midrule
\multicolumn{10}{l}{\textit{World Action Models}} \\
OpenWAM-$\alpha$~\cite{wang2026openwam} & $\checkmark$ & 33.2 & \textbf{75.9} & 87.6 & 97.0 & 85.6 & 39.9 & 77.4 & 68.8 \\
Fast-WAM~\cite{yuan2026fastwam} & $\times$ & 16.4 & 44.5 & 68.9 & 78.2 & 53.7 & 37.7 & 60.7 & 51.5 \\
X-WAM~\cite{guo2026xwam} & $\checkmark$ & 72.9 & 46.3 & 88.5 & 93.6 & 88.3 & 62.9 & 80.8 & 74.8 \\
4D-WAM~\cite{yang20264dwam} & $\checkmark$ & 45.2 & 64.3 & \textbf{90.6} & 94.3 & 57.7 & 69.1 & 79.2 & 71.0 \\
ImageWAM~\cite{imagewam} & $\times$ & 77.7 & 48.3 & \underline{89.9} & \underline{97.5} & 86.0 & \underline{95.2} & 81.8 & 81.6 \\
\rowcolor{oursrow}
\textbf{\method{}} & $\times$ & \textbf{83.1} & 57.0 & \textbf{90.6} & \textbf{98.3} & 88.0 & \textbf{95.8} & \underline{83.0} & \textbf{84.5} \\
\bottomrule
\end{tabularx}
\end{table}

\begin{table}[t]
\centering
\caption{\textbf{RoboTwin 2.0 Clean-to-Random.} Success rates (\%) after training solely on clean demonstrations. P.T.\ denotes embodied-policy pretraining; Geo.\ denotes geometric inputs or supervision.}
\label{tab:comparison-robotwin-c2r}
\footnotesize
\setlength{\tabcolsep}{9pt}
\renewcommand{\arraystretch}{1.14}
\begin{tabularx}{0.76\textwidth}{Xccccc}
\toprule
Method & P.T. & Geo. & Clean & Rand. & Avg. \\
\midrule
\multicolumn{6}{l}{\textit{Vision-Language-Action Models}} \\
$\pi_{0.5}$~\cite{intelligence2025pi05} & $\checkmark$ & $\times$ & 70.66 & 46.00 & 58.33 \\
\midrule
\multicolumn{6}{l}{\textit{World Action Models}} \\
Fast-WAM~\cite{yuan2026fastwam} & $\times$ & $\times$ & 77.76 & 1.92 & 39.84 \\
UniWAM~\cite{chen2026uniwam} & $\checkmark$ & $\times$ & 75.12 & \textbf{67.56} & \underline{71.34} \\
X-WAM~\cite{guo2026xwam} & $\checkmark$ & $\checkmark$ & 69.98 & 25.84 & 47.91 \\
4D-WAM~\cite{yang20264dwam} & $\checkmark$ & $\checkmark$ & 81.50 & 41.80 & 61.65 \\
ImageWAM~\cite{imagewam} & $\checkmark$ & $\times$ & \underline{81.88} & 58.82 & 70.35 \\
\rowcolor{oursrow}
\textbf{\method{}} & $\checkmark$ & $\checkmark$ & \textbf{87.84} & \underline{67.06} & \textbf{77.45} \\
\bottomrule
\end{tabularx}
\end{table}

\subsection{Ablation and Mechanism Analysis}
\label{sec:ablations}
\begin{figure*}[!t]
\centering
\captionsetup{skip=3pt}
\includegraphics[width=\textwidth]{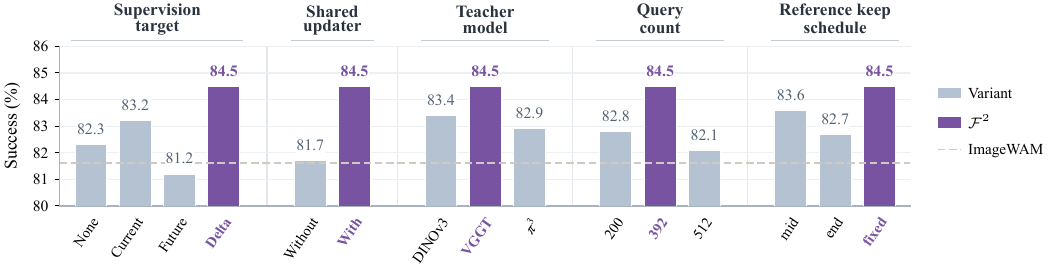}
\caption{\textbf{Ablation study on LIBERO-Plus.} Each group varies one factor while retaining the other default settings; all variants are trained for 10 epochs.}
\label{fig:ablations}
\end{figure*}

We conduct ablation studies on LIBERO-Plus to examine the supervision and design of spatial transition queries. All variants use the same training data as the \method{} baseline and are trained for 10 epochs. Figure~\ref{fig:ablations} summarizes the results, which we discuss through three questions.

\paragraph{Does supervising geometric transitions improve the queries?}
We first examine whether the queries benefit from a teacher target. In the supervision-target comparison in Figure~\ref{fig:ablations}, feature-delta supervision improves success from 82.3\% without distillation to 84.5\%, indicating a contribution beyond the additional query tokens. We then compare what the teacher supplies: the aligned feature difference outperforms both current- and future-feature targets, which reach 83.2\% and 81.2\%, respectively. Unlike an individual-frame target, the feature difference relates the paired observations and supervises how their teacher representations change; however, the target remains a feature-space representation rather than a direct measurement of 3D motion.

\paragraph{How does target construction affect performance?}
We next examine the effect of the teacher representation used to construct the feature-difference target. DINOv3~\cite{simeoni2025dinov3}, VGGT, and $\pi^3$~\cite{wang2025pi3} yield 83.4\%, 84.5\%, and 82.9\%, respectively. Among the tested teachers, VGGT achieves the strongest performance. Its improvement over DINOv3 suggests that geometry-oriented teacher features can provide useful spatial supervision for manipulation, while the result with $\pi^3$ shows that the choice of geometric teacher also matters.

\paragraph{How do query design and action conditioning affect performance?}
Finally, we examine how the transition signal is represented and used for action generation. The frame-matched budget of $N=392$ queries (LIBERO's $14\times 28$ two-view token grid) performs best among the tested sizes, while both 200 and 512 queries perform worse, showing that simply increasing the budget does not guarantee improvement. This comparison varies the query budget and does not isolate the effect of spatial correspondence from other changes in the representation. With the query grid in place, removing the shared updater while retaining native masked attention (leaving $Z_\Delta$ self-attention-only, without access to current visual or instruction tokens) reduces success from 84.5\% to 81.7\%, supporting the complete updater design for integrating current visual and conditioning states. This comparison tests the updater as a whole, including its learnable computation and parameter sharing across blocks. A fixed reference keep probability of $q=0.5$ outperforms schedules that anneal $q$ to zero by the midpoint or end of training, supporting the retention of reference-assisted examples alongside $Z_\Delta$.

\paragraph{Learned transition visualization.}
\begin{figure*}[t]
    \vspace{-5pt}
    \centering
    \captionsetup[subfigure]{position=bottom,font={footnotesize,rm},skip=2pt,justification=centering}
    \begin{subfigure}[t]{0.98\textwidth}
        \centering
        \includegraphics[width=\linewidth]{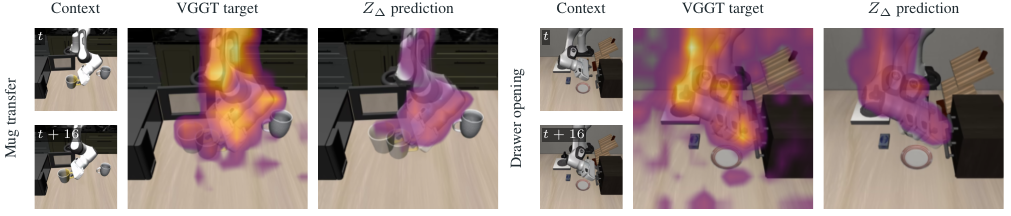}
        \caption{Feature-change visualization.}
        \label{fig:transition-feature-readout}
    \end{subfigure}\par\vspace{2pt}
    \begin{subfigure}[t]{0.98\textwidth}
        \centering
        \includegraphics[width=\linewidth]{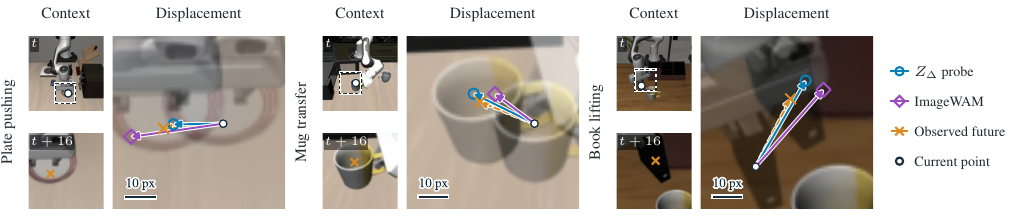}
        \caption{Future-point visualization.}
        \label{fig:transition-point-readout}
    \end{subfigure}
    \caption{\textbf{Learned transition visualization.}
    (a) Color encodes the $\ell_2$ magnitude of VGGT patch-feature differences
    between flow-aligned current/future observations, or the corresponding
    student prediction: purple indicates smaller differences, and
    orange/yellow larger differences.
    (b) Future-point predictions from a supervised probe on frozen $Z_\Delta$
    are compared with tracking references in ImageWAM-generated and observed
    futures.}
    \label{fig:learned-transition-readouts}
    \vspace{-5pt}
\end{figure*}

Figure~\ref{fig:learned-transition-readouts} visualizes information captured by the transition queries. In (a), the predicted feature-change maps highlight the moving arm and nearby interaction regions, overlapping with areas of elevated response in the VGGT targets. In (b), a separately trained displacement probe reads out future point locations from frozen $Z_\Delta$; its predictions lie near the observed-future tracking references in the illustrated examples. Together, these examples suggest that the queries contain spatially localized transition information that can be read out by a trained probe to estimate future image-space displacement. Feature-change and displacement-probe visualization details are provided in Appendix~\ref{sec:implementation}.

\begin{figure}[t]
\vspace{-5pt}
\centering
\includegraphics[width=0.75\columnwidth]{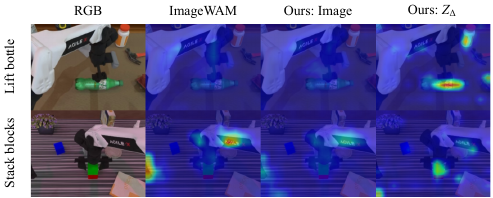}
\caption{\textbf{Attention routing on randomized RoboTwin tasks.} Positive-gradient rollout for the first action's left-gripper output at the final denoising step.}
\label{fig:robotwin-attention}
\vspace{-5pt}
\end{figure}

\paragraph{Action-conditioned attention.}
Figure~\ref{fig:robotwin-attention} compares ImageWAM and \method{} on two randomized RoboTwin observations. The gradient-weighted attention rollout shows responses associated with visual and transition-query tokens for the first left-gripper action. In these examples, the transition-query map highlights the bottle in the lifting task and the manipulated block in the stacking task, while the image-token maps also respond to regions of the robot arm. Compared with ImageWAM, whose responses are less concentrated on the manipulated objects in these examples, the query pathway highlights object and interaction regions during action generation. The attention rollout protocol is described in Appendix~\ref{sec:additional}.

\subsection{Real-World Evaluation}
\label{sec:real-world}
\begin{figure}[tbp]
\centering
\includegraphics[width=\textwidth]{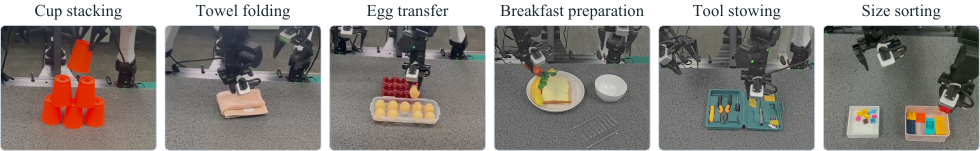}
\captionsetup{skip=5pt}
\caption{\textbf{Real-world tasks.} Representative front-view frames from the six evaluation tasks.}
\label{fig:real-world-overview}
\end{figure}

We evaluate physical manipulation on six tasks: cup stacking, towel folding, egg transfer, breakfast preparation, tool stowing, and size sorting. We compare $\pi_{0.5}$~\cite{intelligence2025pi05}, ImageWAM~\cite{imagewam}, and \method{}, each jointly trained on the six tasks. Table~\ref{tab:real} summarizes the comparison over 15 trials per task, and Figure~\ref{fig:real-world-overview} illustrates the tasks; demonstration statistics and detailed success criteria are provided in Appendix~\ref{sec:real-world-details}.

\paragraph{Success criteria.}
Cup stacking, towel folding, egg transfer, and breakfast preparation require full-task completion. Tool stowing and size sorting use item-level success rates. Each tool-stowing trial contains four tools, while each size-sorting trial contains 15 blocks and is scored by the fraction correctly sorted.

\begin{table}[t]
\centering
\begin{minipage}[t]{0.47\textwidth}
\centering
\begin{minipage}[t][48pt][t]{\linewidth}
\captionof{table}{\textbf{Real-world task scores (\%).} $^*$ marks item-level evaluation. Average is the unweighted mean of the six task scores, computed before rounding.}
\label{tab:real}
\end{minipage}\par\vspace{2pt}
\footnotesize
\setlength{\tabcolsep}{4pt}
\renewcommand{\arraystretch}{1.14}
\begin{tabularx}{\linewidth}{Xccc}
\toprule
Task & $\pi_{0.5}$ & ImageWAM & \textbf{\method{}} \\
\midrule
Cup stacking & \underline{73} & 67 & \textbf{93} \\
Towel folding & \textbf{100} & \underline{87} & \textbf{100} \\
Egg transfer & \textbf{80} & \underline{73} & \textbf{80} \\
Breakfast preparation & \textbf{87} & 73 & \underline{80} \\
Tool stowing$^*$ & \underline{63.3} & 31.7 & \textbf{65.0} \\
Size sorting$^*$ & \textbf{97.8} & 89.3 & \underline{96.4} \\
\midrule
\rowcolor{oursrow}
\textbf{Average} & \underline{83.5} & 70.2 & \textbf{85.8} \\
\bottomrule
\end{tabularx}
\end{minipage}\hfill
\begin{minipage}[t]{0.49\textwidth}
\centering
\begin{minipage}[t][48pt][t]{\linewidth}
\captionof{table}{\textbf{Inference efficiency.} Median CUDA-event latency per 16-action chunk on one H200 (BF16, 10 denoising steps, online text encoding), relative to eager ImageWAM.}
\label{tab:efficiency}
\end{minipage}\par\vspace{2pt}
\footnotesize
\setlength{\tabcolsep}{4pt}
\renewcommand{\arraystretch}{1.28}
\begin{tabularx}{\linewidth}{Xcc}
\toprule
Method & \makecell[c]{Relative\\latency} & \makecell[c]{Latency\\(ms/chunk)} \\
\midrule
ImageWAM~\cite{imagewam} & $1.000\times$ & $187.41\,{\pm}\,4.80$ \\
\rowcolor{oursrow}
\textbf{\method{}} & $1.041\times$ & $195.01\,{\pm}\,2.99$ \\
\midrule
ImageWAM (CUDA Graph) & $0.428\times$ & $80.20\,{\pm}\,3.74$ \\
\rowcolor{oursrow}
\method{} (CUDA Graph) & $0.531\times$ & $99.53\,{\pm}\,3.26$ \\
\bottomrule
\end{tabularx}
\end{minipage}
\end{table}

\method{} achieves an average score of 85.8\%, compared with 83.5\% for $\pi_{0.5}$ and 70.2\% for ImageWAM. Whereas ImageWAM reports strong real-world performance on its original suite~\cite{imagewam}, its lower scores here reflect differences in task suites, demonstrations, and evaluation protocols. In particular, several of our tasks require consecutive manipulation steps, during which the policy must respond to changing object configurations. The learned transition representation may provide useful spatial information for these tasks. Although \method{} is trained without robot pretraining, its results are competitive with the pretrained $\pi_{0.5}$ on several tasks. Further evaluation should focus on generalization.

\subsection{Inference Efficiency}
We assess the inference cost of spatial transition conditioning in Table~\ref{tab:efficiency}. Under the same measurement settings, \method{} takes $195.01\,{\pm}\,2.99$\,ms per 16-action chunk, compared with $187.41\,{\pm}\,4.80$\,ms for ImageWAM. This modest 4.1\% increase accompanies the manipulation improvements reported above. The small overhead indicates that the additional query pathway largely preserves the runtime efficiency of the underlying WAM.

To improve inference speed for real-world deployment, we apply CUDA Graph optimization to both methods. The optimized ImageWAM and \method{} achieve $80.20\,{\pm}\,3.74$\,ms and $99.53\,{\pm}\,3.26$\,ms per chunk, respectively. For \method{}, this reduces the inference latency by approximately 49\%, supporting more responsive action generation during real-world execution.

\section{Conclusion}
\method{} learns spatial transition queries from correspondence-aligned changes in pretrained 3D features. These queries provide the Action Expert with a representation of anticipated spatial change, connecting geometric supervision during training to action generation from current observations. Experiments on LIBERO, LIBERO-Plus, RoboTwin~2.0, and six real-world tasks show strong manipulation performance and improved robustness. Together, the results suggest that supervising how a scene changes in 3D feature space is a useful complement to future-image prediction in world action models.  Studying how spatial transition learning generalizes across broader tasks, embodiments, and environments is an important direction for future research.

\clearpage
\begingroup
  \small
  \bibliographystyle{ieeenat_fullname}
  \bibliography{main}

@article{imagewam,
  title={{ImageWAM}: Do World Action Models Really Need Video Generation, or Just Image Editing?},
  author={Zhang, Yuyang and Zhang, Wenyao and Qi, Zekun and Zhang, He and Lin, Haitao and Zhang, Jingbo and Mu, Yao and Yang, Xiaokang and Zeng, Wenjun and Jin, Xin},
  journal={arXiv preprint arXiv:2606.19531},
  year={2026}
}

@inproceedings{wang2025vggt,
  title={{VGGT}: Visual Geometry Grounded Transformer},
  author={Wang, Jianyuan and Chen, Minghao and Karaev, Nikita and Vedaldi, Andrea and Rupprecht, Christian and Novotny, David},
  booktitle={Proceedings of the IEEE/CVF Conference on Computer Vision and Pattern Recognition (CVPR)},
  pages={5294--5306},
  year={2025}
}

@inproceedings{wu2025geometryforcing,
  title={Geometry Forcing: Marrying Video Diffusion and {3D} Representation for Consistent World Modeling},
  author={Wu, Haoyu and Wu, Diankun and He, Tianyu and Guo, Junliang and Ye, Yang and Duan, Yueqi and Bian, Jiang},
  booktitle={International Conference on Learning Representations},
  year={2026}
}

@inproceedings{LIBERO23,
  title={{LIBERO:} Benchmarking Knowledge Transfer for Lifelong Robot Learning},
  author={Liu, Bo and Zhu, Yifeng and Gao, Chongkai and Feng, Yihao and Liu, Qiang and Zhu, Yuke and Stone, Peter},
  booktitle={Advances in Neural Information Processing Systems},
  volume={36},
  pages={44776--44791},
  year={2023},
  doi={10.52202/075280-1939}
}

@article{wang2026openwam,
  title={{OpenWAM}: An Open, Modular Exploration Towards Systematic World-Action Model Pretraining},
  author={Wang, Yuran and Huang, Siqiao and Li, Mingleyang and Zhang, Chenhao and Liang, Jiaqi and Jin, Weiyang and Chen, Yue and Chi, Xuemin and Zhou, Donghao and Yu, Qize and Wang, Yu-Kai and Rui, Yuhan and Yao, Shenzhe and Yuan, Zhen and Shen, Zhenhao and Zhu, Kefei and Zhu, Zijie and Gao, Ning and Chi, Xiaowei and He, Guanqi and Zhang, Shanghang and Dong, Hao and Shao, Lin and Zhao, Hang},
  journal={arXiv preprint arXiv:2609.07398},
  year={2026}
}

@article{starvla2026,
  title={{StarVLA}: A Lego-like Codebase for Vision-Language-Action Model Developing},
  author={{StarVLA Community}},
  journal={arXiv preprint arXiv:2604.05014},
  year={2026}
}

@article{yang20264dwam,
  title={{4D-WAM}: Infusing Spatiotemporal Awareness into World Action Models through Trajectory Fields},
  author={Yang, Lishan and Song, Wenxuan and Wang, Xi and Sheng, Pingyue and Fang, Zheng and Zhou, Ziyang and He, Junjie and Yan, Haodong and Chen, Jiayi and Sun, Nan and Sun, Qiao and Wang, Pengwei and Liu, Lingqiao and Wang, Yan and Gao, Yuxiang and Dayoub, Feras and Li, Haoang},
  journal={arXiv preprint arXiv:2608.08023},
  year={2026}
}

@inproceedings{fei25libero-plus,
  title={{LIBERO-Plus}: A Progressive Robustness Benchmark for Visual-Language-Action Models},
  author={Fei, Senyu and Wang, Siyin and Shi, Junhao and Dai, Zihao and Cai, Jikun and Qian, Pengfang and Ji, Li and He, Xinzhe and Zhang, Shiduo and Fei, Zhaoye and Fu, Jinlan and Gong, Jingjing and Qiu, Xipeng},
  booktitle={Proceedings of the IEEE/CVF Conference on Computer Vision and Pattern Recognition (CVPR)},
  pages={38574--38583},
  year={2026}
}

@inproceedings{chen2025robotwin,
  title={{RoboTwin} 2.0: A scalable data generator and benchmark with strong domain randomization for robust bimanual robotic manipulation},
  author={Chen, Tianxing and Chen, Zanxin and Chen, Baijun and Cai, Zijian and Liu, Yibin and Li, Zixuan and Liang, Qiwei and Lin, Xianliang and Ge, Yiheng and Gu, Zhenyu and Deng, Weiliang and Guo, Yubin and Nian, Tian and Xie, Xuanbing and Chen, Qiangyu and Su, Kailun and Xu, Tianling and Liu, Guodong and Hu, Mengkang and Gao, Huan-Ang and Wang, Kaixuan and Liang, Zhixuan and Qin, Yusen and Yang, Xiaokang and Luo, Ping and Mu, Yao},
  booktitle={Proceedings of the 43rd International Conference on Machine Learning},
  pages={13673--13699},
  year={2026}
}

@inproceedings{su2026worldguidance,
  title={World Guidance: World Modeling in Condition Space for Action Generation},
  author={Su, Yue and Chen, Sijin and Shi, Haixin and Liu, Mingyu and Zhang, Zhengshen and Huang, Ningyuan and Zhong, Weiheng and Zhu, Zhengbang and Liu, Yuxiao and Liu, Xihui},
  booktitle={Proceedings of the 43rd International Conference on Machine Learning},
  pages={115917--115933},
  year={2026}
}

@article{yuan2026fastwam,
  title={{Fast-WAM}: Do World Action Models Need Test-Time Future Imagination?},
  author={Yuan, Tianyuan and Dong, Zibin and Liu, Yicheng and Zhao, Hang},
  journal={arXiv preprint arXiv:2603.16666},
  year={2026}
}

@inproceedings{kim2025openvlaoft,
  title={Fine-Tuning Vision-Language-Action Models: Optimizing Speed and Success},
  author={Moo Jin Kim and Chelsea Finn and Percy Liang},
  booktitle={Proceedings of Robotics: Science and Systems},
  year={2025},
  doi={10.15607/RSS.2025.XXI.017}
}

@inproceedings{intelligence2025pi05,
  title={{$\pi_{0.5}$}: A Vision-Language-Action Model with Open-World Generalization},
  author={Black, Kevin and Brown, Noah and Darpinian, James and Dhabalia, Karan and Driess, Danny and Esmail, Adnan and Equi, Michael Robert and Finn, Chelsea and Fusai, Niccolo and Galliker, Manuel Y. and Ghosh, Dibya and Groom, Lachy and Hausman, Karol and Ichter, Brian and Jakubczak, Szymon and Jones, Tim and Ke, Liyiming and LeBlanc, Devin and Levine, Sergey and Li-Bell, Adrian and Mothukuri, Mohith and Nair, Suraj and Pertsch, Karl and Ren, Allen Z. and Shi, Lucy Xiaoyang and Smith, Laura and Springenberg, Jost Tobias and Stachowicz, Kyle and Tanner, James and Vuong, Quan and Walke, Homer and Walling, Anna and Wang, Haohuan and Yu, Lili and Zhilinsky, Ury},
  booktitle={Proceedings of The 9th Conference on Robot Learning},
  pages={17--40},
  year={2025}
}

@inproceedings{li2026spatial,
  title={Spatial Forcing: Implicit Spatial Representation Alignment for Vision-Language-Action Model},
  author={Li, Fuhao and Song, Wenxuan and Zhao, Han and Wang, Jingbo and Ding, Pengxiang and Wang, Donglin and Zeng, Long and Li, Haoang},
  year={2026},
  booktitle={International Conference on Learning Representations}
}

@inproceedings{bi2026motus,
  title={Motus: A Unified Latent Action World Model},
  author={Bi, Hongzhe and Tan, Hengkai and Xie, Shenghao and Wang, Zeyuan and Huang, Shuhe and Liu, Haitian and Zhao, Ruowen and Feng, Yao and Xiang, Chendong and Rong, Yinze and Zhao, Hongyan and Liu, Hanyu and Su, Zhizhong and Ma, Lei and Su, Hang and Zhu, Jun},
  booktitle={Proceedings of the IEEE/CVF Conference on Computer Vision and Pattern Recognition (CVPR)},
  pages={35101--35113},
  year={2026}
}

@inproceedings{lingbotva,
  title={Causal World Modeling for Robot Control},
  author={Lin Li and Qihang Zhang and Yiming Luo and Shuai Yang and Ruilin Wang and Luyao Zhang and Mingrui Yu and Zelin Gao and Nan Xue and Boyu Zhou and Xing Zhu and Mingyu Ding and Yujun Shen and Yinghao Xu},
  booktitle={Proceedings of Robotics: Science and Systems},
  year={2026},
  doi={10.15607/RSS.2026.XXII.016}
}

@article{guo2026xwam,
  title={Unified {4D} World Action Modeling from Video Priors with Asynchronous Denoising},
  author={Guo, Jun and Li, Qiwei and Li, Peiyan and Chen, Zilong and Sun, Nan and Su, Yifei and Wang, Heyun and Zhang, Yuan and Li, Xinghang and Liu, Huaping},
  journal={arXiv preprint arXiv:2604.26694},
  year={2026}
}

@article{li2026wam4d,
  title={{WAM4D}: Fast {4D} World Action Model via Spatial Register Tokens},
  author={Li, Ying and Wei, Xiaobao and Cao, Jiajun and Wang, Hao and Chi, Xiaowei and Bai, Chengyu and Sun, Qianpu and Li, Jiajun and Zhang, Xiaojie and Jia, Peidong and Tang, Jian and Han, Sirui and Zhang, Shanghang},
  journal={arXiv preprint arXiv:2606.14048},
  year={2026}
}

@article{ma2026geosemwam,
  title={{GeoSem-WAM}: Geometry- and Semantic-Aware World Action Models},
  author={Ma, Fulong and Peng, Daojie and Yue, Wenjun and Cao, Jiahang and Wang, Bintao and Zhang, Qiang and Ma, Jun},
  journal={arXiv preprint arXiv:2606.03188},
  year={2026}
}

@article{zhang2026mecowam,
  title={Learning {4D} Geometric Priors for Inference-Efficient World Action Models},
  author={Zhang, Jianjun and Zhu, Jian and Su, Taiyi and Ma, Chong and Huang, Zitai and Xu, Yi and Wang, Hanli},
  journal={arXiv preprint arXiv:2607.05468},
  year={2026}
}

@inproceedings{24pi0,
  title={{$\pi_0$}: A Vision-Language-Action Flow Model for General Robot Control},
  author={Kevin Black and Noah Brown and Danny Driess and Adnan Esmail and Michael Robert Equi and Chelsea Finn and Niccolo Fusai and Lachy Groom and Karol Hausman and Brian Ichter and Szymon Jakubczak and Tim Jones and Liyiming Ke and Sergey Levine and Adrian Li-Bell and Mohith Mothukuri and Suraj Nair and Karl Pertsch and Lucy Xiaoyang Shi and Laura Smith and James Tanner and Quan Vuong and Anna Walling and Haohuan Wang and Ury Zhilinsky},
  booktitle={Proceedings of Robotics: Science and Systems},
  year={2025},
  doi={10.15607/RSS.2025.XXI.010}
}

@article{cen2025worldvla,
  title={{WorldVLA}: Towards Autoregressive Action World Model},
  author={Cen, Jun and Yu, Chaohui and Yuan, Hangjie and Jiang, Yuming and Huang, Siteng and Guo, Jiayan and Li, Xin and Song, Yibing and Luo, Hao and Wang, Fan and Zhao, Deli and Chen, Hao},
  journal={arXiv preprint arXiv:2506.21539},
  year={2025}
}

@article{huang2026pointworld,
  title={{PointWorld}: Scaling {3D} World Models for In-The-Wild Robotic Manipulation},
  author={Huang, Wenlong and Chao, Yu-Wei and Mousavian, Arsalan and Liu, Ming-Yu and Fox, Dieter and Mo, Kaichun and Fei-Fei, Li},
  journal={arXiv preprint arXiv:2601.03782},
  year={2026}
}

@article{yan2026flexpi,
  title={Flex-$\pi$: A Multi-Stream World-Action Model with Compute Flexibility},
  author={Yan, Ge and Liu, Jinghao and Fan, Yuzhi and Cai, Lei and Liao, Minwen and Zhang, Jesse and Fox, Dieter},
  journal={arXiv preprint arXiv:2608.10860},
  year={2026}
}

@article{park2026pointwam,
  title={{PointWAM}: {3D} World Action Modeling for Dexterous Robotic Manipulation},
  author={Park, Chunghyun and Kim, Beomjun and Park, Seungcheol and Kwon, Heeseung and Shukla, Yashu and Sim, Seunghoon and Shin, Jinwoo and Cho, Minsu},
  journal={arXiv preprint arXiv:2610.02840},
  year={2026}
}

@article{chen2026uniwam,
  title={{UniWAM}: Unified World-Action Model},
  author={Song, Wenxuan and Chen, Jiayi and Wang, Jingbo and Zhou, Shuai and Gong, Xicheng and Fan, Zehua and Zhou, Ziyang and E, Junwu and Yan, Haodong and Li, Fuhao and Yu, Qize and Huang, Xu and Wang, Pengwei and Chen, Wen and Zhou, Shunbo and Li, Haoang},
  journal={arXiv preprint arXiv:2610.02054},
  year={2026}
}

@inproceedings{wang2026vggtomega,
  title={{VGGT-$\Omega$}},
  author={Wang, Jianyuan and Chen, Minghao and Zhang, Shangzhan and Karaev, Nikita and Sch\"onberger, Johannes and Labatut, Patrick and Bojanowski, Piotr and Novotny, David and Vedaldi, Andrea and Rupprecht, Christian},
  booktitle={Proceedings of the IEEE/CVF Conference on Computer Vision and Pattern Recognition (CVPR)},
  pages={21486--21499},
  year={2026}
}

@article{simeoni2025dinov3,
  title={{DINOv3}},
  author={Sim{\'{e}}oni, Oriane and Vo, Huy V. and Seitzer, Maximilian and Baldassarre, Federico and Oquab, Maxime and Jose, Cijo and Khalidov, Vasil and Szafraniec, Marc and Yi, Seungeun and Ramamonjisoa, Micha{\"{e}}l and Massa, Francisco and Haziza, Daniel and Wehrstedt, Luca and Wang, Jianyuan and Darcet, Timoth{\'{e}}e and Moutakanni, Th{\'{e}}o and Sentana, Leonel and Roberts, Claire and Vedaldi, Andrea and Tolan, Jamie and Brandt, John and Couprie, Camille and Mairal, Julien and J{\'{e}}gou, Herv{\'{e}} and Labatut, Patrick and Bojanowski, Piotr},
  journal={Transactions on Machine Learning Research},
  year={2026},
  url={https://openreview.net/forum?id=2NlGyqNjns}
}

@inproceedings{wang2025pi3,
  title={{$\pi^3$}: Permutation-Equivariant Visual Geometry Learning},
  author={Wang, Yifan and Zhou, Jianjun and Zhu, Haoyi and Chang, Wenzheng and Zhou, Yang and Li, Zizun and Chen, Junyi and Pang, Jiangmiao and Shen, Chunhua and He, Tong},
  year={2026},
  booktitle={International Conference on Learning Representations}
}

@inproceedings{abnar2020quantifying,
  title={Quantifying Attention Flow in Transformers},
  author={Abnar, Samira and Zuidema, Willem},
  booktitle={Proceedings of the 58th Annual Meeting of the Association for Computational Linguistics},
  pages={4190--4197},
  publisher={Association for Computational Linguistics},
  year={2020},
  doi={10.18653/v1/2020.acl-main.385},
  url={https://aclanthology.org/2020.acl-main.385/}
}

@inproceedings{karaev2024cotracker3,
  title={{CoTracker3}: Simpler and Better Point Tracking by Pseudo-Labelling Real Videos},
  author={Karaev, Nikita and Makarov, Yuri and Wang, Jianyuan and Neverova, Natalia and Vedaldi, Andrea and Rupprecht, Christian},
  booktitle={Proceedings of the IEEE/CVF International Conference on Computer Vision (ICCV)},
  pages={6013--6022},
  year={2025}
}

@article{wang2026unitas,
  title={{UNITAS}: A {3D}-Native World Action Model for Embodied Manipulation},
  author={Wang, Ruixiang and Su, Yongyi and Zhou, Wenlve and Yue, Bo and Liu, Hengyan and Lu, Dekun and Tian, Yuxin and Fang, Yihan and Wu, Zerui and Hu, Xing and Chen, Jietao and Guo, Yong and He, Ziyan and Yuan, Junbin and Liu, Guiliang and Xing, Xiaofen and Jia, Kui},
  journal={arXiv preprint arXiv:2610.12099},
  year={2026}
}

@article{chen2026mvgwam,
  title={{MVG-WAM}: Multiple View Geometry-Aware World-Action Modeling for Robotic Manipulation},
  author={Chen, Wenbo and Li, Tianfu and Xu, Haoxuan and Cao, Zhihao and Chen, Zhenghan and Zhu, Zhengming and Luo, Zizhou and Yang, Guosheng and Liu, Yuan and Wang, Lujia and Chen, Wen and Li, Haoang},
  journal={arXiv preprint arXiv:2609.37793},
  year={2026}
}

@inproceedings{yang2025fast3r,
  title={{Fast3R}: Towards {3D} Reconstruction of 1000+ Images in One Forward Pass},
  author={Yang, Jianing and Sax, Alexander and Liang, Kevin J. and Henaff, Mikael and Tang, Hao and Cao, Ang and Chai, Joyce and Meier, Franziska and Feiszli, Matt},
  booktitle={Proceedings of the IEEE/CVF Conference on Computer Vision and Pattern Recognition (CVPR)},
  pages={21924--21935},
  year={2025}
}

@inproceedings{wang2025cut3r,
  title={Continuous {3D} Perception Model with Persistent State},
  author={Wang, Qianqian and Zhang, Yifei and Holynski, Aleksander and Efros, Alexei A. and Kanazawa, Angjoo},
  booktitle={Proceedings of the IEEE/CVF Conference on Computer Vision and Pattern Recognition (CVPR)},
  pages={10510--10522},
  year={2025}
}

@inproceedings{lin2025da3,
  title={{Depth Anything 3}: Recovering the Visual Space from Any Views},
  author={Lin, Haotong and Chen, Sili and Liew, Jun Hao and Chen, Donny Y. and Li, Zhenyu and Zhao, Yang and Peng, Sida and Guo, Hengkai and Zhou, Xiaowei and Shi, Guang and Feng, Jiashi and Kang, Bingyi},
  booktitle={International Conference on Learning Representations},
  year={2026},
  pages={141261--141285}
}

@inproceedings{leroy2024mast3r,
  title={Grounding Image Matching in {3D} with {MASt3R}},
  author={Leroy, Vincent and Cabon, Yohann and Revaud, J{\'e}r{\^o}me},
  booktitle={European Conference on Computer Vision},
  year={2024},
  pages={71--91},
  doi={10.1007/978-3-031-73220-1_5}
}

@inproceedings{wang2024dust3r,
  title={{DUSt3R}: Geometric {3D} Vision Made Easy},
  author={Wang, Shuzhe and Leroy, Vincent and Cabon, Yohann and Chidlovskii, Boris and Revaud, Jerome},
  booktitle={Proceedings of the IEEE/CVF Conference on Computer Vision and Pattern Recognition (CVPR)},
  pages={20697--20709},
  year={2024}
}

@inproceedings{ma2026groupediting,
  title={Group Editing: Edit Multiple Images in One Go},
  author={Ma, Yue and Wang, Xinyu and Ma, Qianli and Wang, Qinghe and Zheng, Mingzhe and Yang, Xiangpeng and Li, Hao and Zhao, Chongbo and Ying, Jixuan and Yang, Harry and Liu, Hongyu and Chen, Qifeng},
  booktitle={Proceedings of the IEEE/CVF Conference on Computer Vision and Pattern Recognition (CVPR)},
  pages={43418--43428},
  year={2026}
}

@inproceedings{liu2026camgeo,
  title={{CamGeo}: Sparse Camera-Conditioned Image-to-Video Generation with {3D} Geometry Priors},
  author={Liu, Xuanyi and Ji, Deyi and Liu, Liqun and Zhu, Lanyun and Chen, Xuhang and Xu, Qianxiong and Shu, Peng and Yu, Huan and Jiang, Jie and Gao, Feng and Ma, Siwei},
  booktitle={Proceedings of the 43rd International Conference on Machine Learning},
  pages={76814--76831},
  year={2026}
}

@inproceedings{zitkovich2023rt2,
  title={{RT-2}: Vision-Language-Action Models Transfer Web Knowledge to Robotic Control},
  author={Zitkovich, Brianna and Yu, Tianhe and Xu, Sichun and Xu, Peng and Xiao, Ted and Xia, Fei and Wu, Jialin and Wohlhart, Paul and Welker, Stefan and Wahid, Ayzaan and Vuong, Quan and Vanhoucke, Vincent and Tran, Huong and Soricut, Radu and Singh, Anikait and Singh, Jaspiar and Sermanet, Pierre and Sanketi, Pannag R. and Salazar, Grecia and Ryoo, Michael S. and Reymann, Krista and Rao, Kanishka and Pertsch, Karl and Mordatch, Igor and Michalewski, Henryk and Lu, Yao and Levine, Sergey and Lee, Lisa and Lee, Tsang-Wei Edward and Leal, Isabel and Kuang, Yuheng and Kalashnikov, Dmitry and Julian, Ryan and Joshi, Nikhil J. and Irpan, Alex and Ichter, Brian and Hsu, Jasmine and Herzog, Alexander and Hausman, Karol and Gopalakrishnan, Keerthana and Fu, Chuyuan and Florence, Pete and Finn, Chelsea and Dubey, Kumar Avinava and Driess, Danny and Ding, Tianli and Choromanski, Krzysztof Marcin and Chen, Xi and Chebotar, Yevgen and Carbajal, Justice and Brown, Noah and Brohan, Anthony and Arenas, Montserrat Gonzalez and Han, Kehang},
  booktitle={Proceedings of The 7th Conference on Robot Learning},
  pages={2165--2183},
  year={2023}
}

@inproceedings{kim2025openvla,
  title={{OpenVLA}: An Open-Source Vision-Language-Action Model},
  author={Kim, Moo Jin and Pertsch, Karl and Karamcheti, Siddharth and Xiao, Ted and Balakrishna, Ashwin and Nair, Suraj and Rafailov, Rafael and Foster, Ethan P and Sanketi, Pannag R and Vuong, Quan and Kollar, Thomas and Burchfiel, Benjamin and Tedrake, Russ and Sadigh, Dorsa and Levine, Sergey and Liang, Percy and Finn, Chelsea},
  booktitle={Proceedings of The 8th Conference on Robot Learning},
  pages={2679--2713},
  year={2025}
}

@inproceedings{chefer2021generic,
  title={Generic Attention-Model Explainability for Interpreting Bi-Modal and Encoder-Decoder Transformers},
  author={Chefer, Hila and Gur, Shir and Wolf, Lior},
  booktitle={Proceedings of the IEEE/CVF International Conference on Computer Vision (ICCV)},
  pages={397--406},
  year={2021}
}

@inproceedings{yu2025repa,
  title={Representation Alignment for Generation: Training Diffusion Transformers Is Easier Than You Think},
  author={Sihyun Yu and Sangkyung Kwak and Huiwon Jang and Jongheon Jeong and Jonathan Huang and Jinwoo Shin and Saining Xie},
  year={2025},
  booktitle={International Conference on Learning Representations}
}

@inproceedings{lipman2023flow,
  title={Flow Matching for Generative Modeling},
  author={Lipman, Yaron and Chen, Ricky T. Q. and Ben-Hamu, Heli and Nickel, Maximilian and Le, Matt},
  booktitle={International Conference on Learning Representations},
  year={2023}
}

@article{tian2025interndata,
  title={{InternData-A1}: Pioneering High-Fidelity Synthetic Data for Pre-training Generalist Policy},
  author={Tian, Yang and Yang, Yuyin and Xie, Yiman and Cai, Zetao and Shi, Xu and Gao, Ning and Liu, Hangxu and Jiang, Xuekun and Qiu, Zherui and Yuan, Feng and Li, Yaping and Wang, Ping and Cai, Junhao and Zeng, Jia and Dong, Hao and Pang, Jiangmiao},
  journal={arXiv preprint arXiv:2511.16651},
  year={2025}
}

@misc{flux-2-2025,
  title={{FLUX.2: Frontier Visual Intelligence}},
  author={{Black Forest Labs}},
  howpublished={\url{https://bfl.ai/blog/flux-2}},
  year={2025}
}

@article{wang2026patchwam,
  title={An Action Is Worth One Patch: Unified World-Action Modeling with {PatchWAM}},
  author={Wang, Tianheng and Xie, Zhou and Jia, Heng and Xu, Jianhua and Zhang, Tong and Yu, Kaicheng},
  journal={arXiv preprint arXiv:2609.25961},
  year={2026}
}
\endgroup

\clearpage
\appendix
\small
\setlength{\parskip}{0.32em}
\setlength{\intextsep}{10pt plus 2pt minus 2pt}
\captionsetup{font=footnotesize}
\section{Implementation Details}
\label{sec:implementation}
This appendix provides architectural and optimization configurations, distillation details, per-task RoboTwin~2.0 results, real-world evaluation protocols, attention diagnostics, and baseline evaluation notes.

\subsection{Soft Spatial Alignment}
\label{sec:appendix-flow}
For each view $v$, we estimate dense backward optical flow $f^v$ (target-to-reference Farneb\"ack flow) from the future observation $I_{t+H}^v$ to the current observation $I_t^v$, and warp the current image onto the future-image grid:
\begin{equation}
    \widetilde I_t^v(x)=I_t^v\!\left(x+f^v(x)\right),
    \label{eq:warp}
\end{equation}
where $x$ indexes a pixel on the future grid. A binary validity map records whether each warped coordinate falls inside the current image bounds and is average-pooled to obtain patch-level validity weights $m_p\in[0,1]$. Alignment is applied prior to frozen VGGT feature extraction. On LIBERO-Plus, enabling soft spatial alignment improves overall success from 81.9\% to 84.5\% (+2.6 percentage points; Table~\ref{tab:alignment-ablation}).

\begin{table}[H]
\centering
\caption{\textbf{Soft spatial alignment ablation on LIBERO-Plus.} Aligning paired observations before VGGT feature extraction yields a $+2.6$ percentage-point gain.}
\label{tab:alignment-ablation}
\footnotesize
\setlength{\tabcolsep}{8pt}
\renewcommand{\arraystretch}{1.14}
\begin{tabularx}{0.48\textwidth}{Xc}
\toprule
Image alignment & Success (\%) \\
\midrule
Without alignment & 81.9 \\
\rowcolor{oursrow}
\textbf{With alignment (\method{})} & \textbf{84.5} \\
\bottomrule
\end{tabularx}
\end{table}

\subsection{Training Configurations}
\label{sec:training-config}
Tables~\ref{table:training_hyperparams}--\ref{table:training_cost} summarize the shared optimization settings, geometry-module hyperparameters, dataset-specific configurations, and wall-clock training costs across all benchmarks.

\begin{table}[H]
\centering
\begin{minipage}[t]{0.485\textwidth}
\centering
\begin{minipage}[t][26pt][t]{\linewidth}
\captionof{table}{\textbf{Common training hyperparameters.} Shared optimization settings across simulation and real-world runs.}
\label{table:training_hyperparams}
\end{minipage}\par\vspace{2pt}
\footnotesize
\setlength{\tabcolsep}{4pt}
\renewcommand{\arraystretch}{1.14}
\begin{tabularx}{\linewidth}{@{}>{\raggedright\arraybackslash}X>{\raggedleft\arraybackslash}X@{}}
\toprule
Parameter & Value \\
\midrule
\addlinespace[2pt]
Training hardware & \makecell[r]{8 or 32 NVIDIA H200\\128 T-Head PPU-ZW810E} \\
\addlinespace[1pt]
Distributed strategy & DeepSpeed ZeRO-1 \\
Precision & BF16 \\
Optimizer & AdamW \\
Optimizer betas & $(0.9, 0.95)$ \\
Learning rate & $1 \times 10^{-4}$ \\
Weight decay & $1 \times 10^{-2}$ \\
LR scheduler & Warmup cosine \\
Warmup steps & $0.05\,T_{\mathrm{total}}$ \\
Minimum LR & $0.01 \times \mathrm{lr}$ \\
Gradient clipping & 1.0 \\
\bottomrule
\end{tabularx}
\end{minipage}\hfill
\begin{minipage}[t]{0.485\textwidth}
\centering
\begin{minipage}[t][26pt][t]{\linewidth}
\captionof{table}{\textbf{Geometry-module hyperparameters.} Settings used for spatial transition distillation and query updating.}
\label{tab:defaults}
\end{minipage}\par\vspace{2pt}
\footnotesize
\setlength{\tabcolsep}{4pt}
\renewcommand{\arraystretch}{1.315}
\begin{tabularx}{\linewidth}{Xr}
\toprule
Parameter & Value \\
\midrule
Reference keep probability $q$ & 0.5 \\
Geometry warm-up $K_w$ & $0.05\,T_{\mathrm{total}}$ \\
Initial warm-up multiplier $\rho_0$ & 0.01 \\
Loss weights $\lambda_{\mathrm{img}},\,\lambda_{\mathrm{act}}$ & 0.5,\,1.0 \\
Geometry loss weight $\lambda_{\mathrm{geo}}$ & 0.1 \\
Alignment weights $\lambda_{\mathrm{ang}},\,\lambda_{\mathrm{scale}}$ & 0.5,\,0.05 \\
Norm regularizer weight $\lambda_{\mathrm{rad}}$ & 0.01 \\
Uniform weight mixture $\eta$ & 0.25 \\
Projector / scale-head width & 1024 / 256 \\
Updater attention heads & 8 \\
\bottomrule
\end{tabularx}
\end{minipage}
\end{table}

\begin{table}[H]
\centering
\caption{\textbf{Dataset-specific training configurations.} Camera-view layouts and image preprocessing follow ImageWAM~\cite{imagewam}; training epochs reflect our evaluation settings.}
\label{table:dataset_training_config}
\footnotesize
\setlength{\tabcolsep}{6pt}
\renewcommand{\arraystretch}{1.16}
\begin{tabular*}{0.82\textwidth}{@{\extracolsep{\fill}}lccc@{}}
\toprule
Parameter & LIBERO & RoboTwin-Full & Real-World \\
\midrule
Input views & 2 & 3 & 3 \\
View layout & Horizontal & \makecell[c]{Wrist-horizontal\\+ vertical} & \makecell[c]{Wrist-horizontal\\+ vertical} \\
Input resolution & $224 \times 448$ & $288 \times 256$ & $288 \times 256$ \\
Future horizon $H$ & 16 frames & 16 frames & 16 frames \\
Action chunk length & 16 & 16 & 16 \\
Training epochs & 10 & 10 & 10 \\
\bottomrule
\end{tabular*}
\end{table}

\begin{table}[H]
\centering
\caption{\textbf{Training time and hardware configuration.} BS denotes batch size per device; Accum.\ denotes gradient accumulation steps.}
\label{table:training_cost}
\footnotesize
\setlength{\tabcolsep}{6pt}
\renewcommand{\arraystretch}{1.14}
\begin{tabular*}{0.82\textwidth}{@{\extracolsep{\fill}}lcr@{}}
\toprule
Benchmark & Training time & $\mathrm{BS} \times \mathrm{Accum.} \times \mathrm{Devices}$ \\
\midrule
LIBERO & 26 hours & $15 \times 1 \times 8$ NVIDIA H200 \\
RoboTwin-Full & 6 days & $24 \times 1 \times 32$ NVIDIA H200 \\
Real-World & 13 hours & $8 \times 1 \times 128$ T-Head PPU-ZW810E \\
\bottomrule
\end{tabular*}
\end{table}

\subsection{Teacher and Spatial Layout}
The frozen VGGT aggregator outputs patch features of channel width $D=2048$. Camera and register tokens are stripped prior to bilinear interpolation onto each camera's region in the student token grid. Our layout supports horizontal and vertical multi-view concatenations as well as the RoboTwin three-view mosaic (a top head view above two side-by-side wrist views). Image crops, teacher features, flow-validity masks (Sec.~\ref{sec:appendix-flow}), and transition queries share identical spatial bounding boxes within the mosaic. Feature subtraction and geometry-loss reductions are computed in FP32 to preserve small feature differences across large teacher activations.

\subsection{Distillation Details}
\label{sec:appendix-loss}
The spatial projector $P_\theta$ maps the transition query grid from width $d$ to the teacher feature width $D$ via two $3\times3$ convolutional layers (each followed by a SiLU activation) and a $1\times1$ output convolution. The scale head $S_\theta$ is a two-layer MLP with an intermediate SiLU activation.

Omitting the batch index for clarity, patch sums and means are computed per training example and subsequently averaged over the batch. The patch weights used in Sec.~\ref{sec:queries} are defined as
\begin{equation}
 w_p=m_p\left[\eta+(1-\eta)
 \frac{\|y_{\Delta,p}\|_2}
 {N^{-1}\sum_{r=1}^{N}\|y_{\Delta,r}\|_2+\epsilon}\right],
 \label{eq:weights}
\end{equation}
where $m_p\in[0,1]$ is the pooled flow-validity mask, $\eta$ balances uniform and magnitude-weighted supervision, $\epsilon$ is a small constant for numerical stability, and the denominator averages over all $N$ student-grid locations across views. The feature-norm regularizer employs one detached target norm per example:
\begin{align}
 r_* &= \max\!\left(
 \frac{\sum_p w_p\|y_{\Delta,p}\|_2}{W},\,1\right),\\
 \mathcal L_{\mathrm{rad}} &= \frac{1}{W}\sum_p w_p\,
 \operatorname{SmoothL1}\!\left(\frac{\|q_p\|_2}{r_*},\,1\right).
\end{align}
This term encourages $\|q_p\|_2\approx r_*$ to stabilize the overall feature scale of the projector, while $S_\theta$ reconstructs patch-specific target features from normalized queries $\widehat q_p$. All teacher-derived weights and norm targets are detached from backpropagation. The geometry warm-up schedule follows $\rho(k)=\rho_0+(1-\rho_0)\min(k/K_w,\,1)$, where $k$ is the optimization step, $\rho_0$ is the initial multiplier, and $K_w = 0.05\,T_{\mathrm{total}}$ is the warm-up horizon relative to the total number of optimization steps $T_{\mathrm{total}}$.

\subsection{Learned Transition Visualization}
\label{sec:transition-visualization-protocol}

\paragraph{Feature-change maps.}
For Figure~\ref{fig:learned-transition-readouts}(a), we compute the channel-wise $\ell_2$ norm of the aligned VGGT feature difference alongside the corresponding prediction from the trained geometry head. External-camera maps are bilinearly upsampled from the $14\times14$ patch grid and rendered under a shared color scale and response threshold to emphasize salient regions. Both maps are overlaid on an equal blend of the current and observed-future frames separated by $H=16$ steps. Future frames are used only to compute the teacher target and visualization background; student queries are predicted solely from the current observation and language instruction.

\paragraph{Future-point visualization.}
For Figure~\ref{fig:learned-transition-readouts}(b), we freeze the trained policy and fit a lightweight two-layer MLP probe to automatically tracked point correspondences. Taking local and pooled $Z_\Delta$ features together with the current query coordinates as input, the probe predicts 2D image-space displacements and is used strictly for post-hoc visualization. Query points are sampled within visible object regions in the current frame, and CoTracker3~\cite{karaev2024cotracker3} tracks them across observed video clips as well as current/ImageWAM-generated image pairs. We retain tracks that remain visible within image bounds and satisfy a 2-pixel forward--backward cycle consistency check. Observed future frames are withheld from both probe prediction and generated-image localization. The plate example is drawn from the held-out probe test split, whereas the mug and book examples come from its training split; CoTracker3 trajectories serve as estimated correspondence references.

\subsection{Data Augmentation}
\label{sec:appendix-augmentation}
During training, we apply synchronized image augmentations across frames within each camera view, including resized cropping, rotation, color jitter, gamma adjustment, and Gaussian noise. Under the RoboTwin Clean-to-Random setting, following PatchWAM~\cite{wang2026patchwam}, we additionally apply stronger appearance augmentations comprising photometric jitter, Gaussian blur, AdaIN-style feature-statistic perturbation, and Fourier amplitude modulation. Spatial transformations are shared between student and teacher inputs so geometric correspondence is preserved, whereas appearance augmentations are applied solely to the student inputs.

\subsection{Training--Inference Visibility}
\label{sec:appendix-visibility}
\begin{figure}[htbp]
    \centering
    \includegraphics[width=0.55\textwidth,clip]{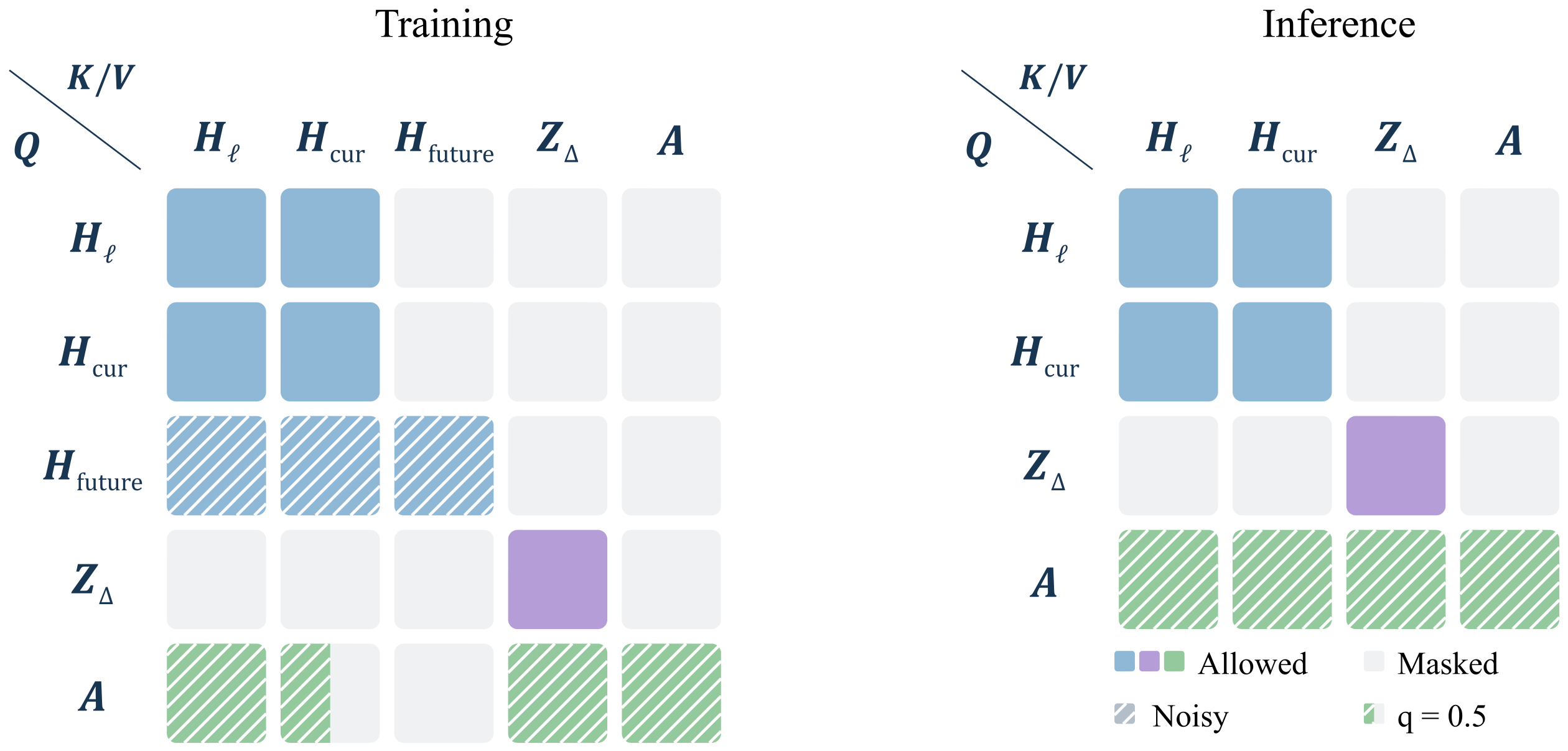}
    \caption{\textbf{Attention visibility.} Native MoT attention masks during training (left) and inference (right), with query rows and key/value columns. The conditioning stream $H_\ell$ combines language instructions and encoded proprioceptive state, and hatched cells denote rows whose query tokens are noisy ($H_{\mathrm{future}}$ and $A$). For the action stream $A$, the half-filled $H_{\mathrm{cur}}$ cell indicates stochastic reference access ($q=0.5$) during training, which becomes full visibility at inference when future tokens are absent. Outside this mask, the shared updater in Figure~\ref{fig:overview} supplies $Z_\Delta$ with current visual and conditioning features.}
    \label{fig:attention-mask}
\end{figure}

Figure~\ref{fig:attention-mask} illustrates the native MoT attention masks used during training and inference. Within the native MoT blocks, transition queries $Z_\Delta$ perform self-attention only, preventing any leakage from future-image tokens during training; they instead aggregate current visual and conditioning features through the shared updater $U_\theta$ using identical current-only context at both training and test time. At inference, future-image tokens are absent and the Action Expert attends jointly to current visual states $H_{\mathrm{cur}}$ and transition queries $Z_\Delta$.

\section{RoboTwin 2.0 Evaluation Details}
\label{sec:robotwin-per-task}

\subsection{Per-Task Results}
Tables~\ref{tab:robotwin-full-detail} and~\ref{tab:robotwin-c2r-detail} (placed at the end of the appendix) report per-task success rates for all 50 tasks in the RoboTwin~2.0 Full and Clean-to-Random benchmarks summarized in Tables~\ref{tab:comparison-robotwin} and~\ref{tab:comparison-robotwin-c2r}, respectively. For \method{}, every task is evaluated over 100 episodes under both clean and randomized conditions; baseline numbers in the Full table follow the ImageWAM codebase.

\subsection{Additional Clean-to-Random Experiments}
\label{sec:c2r-training-controls}
All runs start from the InternData-A1-pretrained ImageWAM model and use the same clean RoboTwin demonstrations and 50-task C2R benchmark. The ImageWAM baseline is trained for 50 epochs without additional appearance augmentation and obtains 70.35\% average success. Our main result trains \method{} for the same 50-epoch stage, followed by 20 additional epochs with additional augmentation, reaching 77.45\%. All variants use the same evaluation protocol, with 100 episodes per task under both clean and randomized conditions.

\paragraph{ImageWAM with additional augmentation.}
Continuing the 50-epoch ImageWAM baseline for 20 more epochs with additional augmentation yields 84.18\% on clean scenes, 67.82\% on randomized scenes, and 76.00\% on average.

\paragraph{\method{} without additional augmentation.}
Continuing the same 50-epoch ImageWAM baseline for 20 more epochs with \method{} but without additional augmentation yields 82.32\%, 60.04\%, and 71.18\% on clean, randomized, and average success, respectively.

Table~\ref{tab:c2r-training-controls} summarizes the training variants and results. Additional augmentation improves the ImageWAM baseline from 70.35\% to 76.00\%, while \method{} further improves the augmented result to 77.45\%. Without augmentation, \method{} reaches 71.18\%, exceeding the ImageWAM baseline. The 0.83-point improvement over ImageWAM without additional augmentation supports the benefit of introducing transition supervision. Under the augmented schedule, \method{} also outperforms ImageWAM by 1.45 points, providing further evidence of its contribution. These results support the value of transition supervision both with and without additional augmentation.

\begin{table}[H]
\centering
\footnotesize
\caption{\textbf{RoboTwin Clean-to-Random training controls.} All runs start from the InternData-A1-pretrained ImageWAM model and use the same clean demonstrations. The reported score is the average of clean and randomized success rates across all 50 tasks.}
\label{tab:c2r-training-controls}
\setlength{\tabcolsep}{5pt}
\renewcommand{\arraystretch}{1.1}
\begin{tabularx}{\textwidth}{l@{\hspace{12pt}}>{\raggedright\arraybackslash}Xccc}
\toprule
Variant & Training schedule & Clean & Rand. & Avg. \\
\midrule
ImageWAM & ImageWAM, 50 ep. & 81.88 & 58.82 & 70.35 \\
\rowcolor{oursrow}
\method{} & \method{}, 50 ep. $+$ \method{}, 20 ep. ($\mathrm{aug.}$) & \textbf{87.84} & \underline{67.06} & \textbf{77.45} \\
ImageWAM w/ aug. & ImageWAM, 50 ep. $+$ ImageWAM, 20 ep. ($\mathrm{aug.}$) & \underline{84.18} & \textbf{67.82} & \underline{76.00} \\
\method{} w/o aug. & ImageWAM, 50 ep. $+$ \method{}, 20 ep. ($\mathrm{w/o\ aug.}$) & 82.32 & 60.04 & 71.18 \\
\bottomrule
\end{tabularx}
\end{table}

\section{Real-World Experimental Details}
\label{sec:real-world-details}

\subsection{Setup and Evaluation Protocol}
\begin{figure*}[p]
    \centering
    \includegraphics[width=\textwidth]{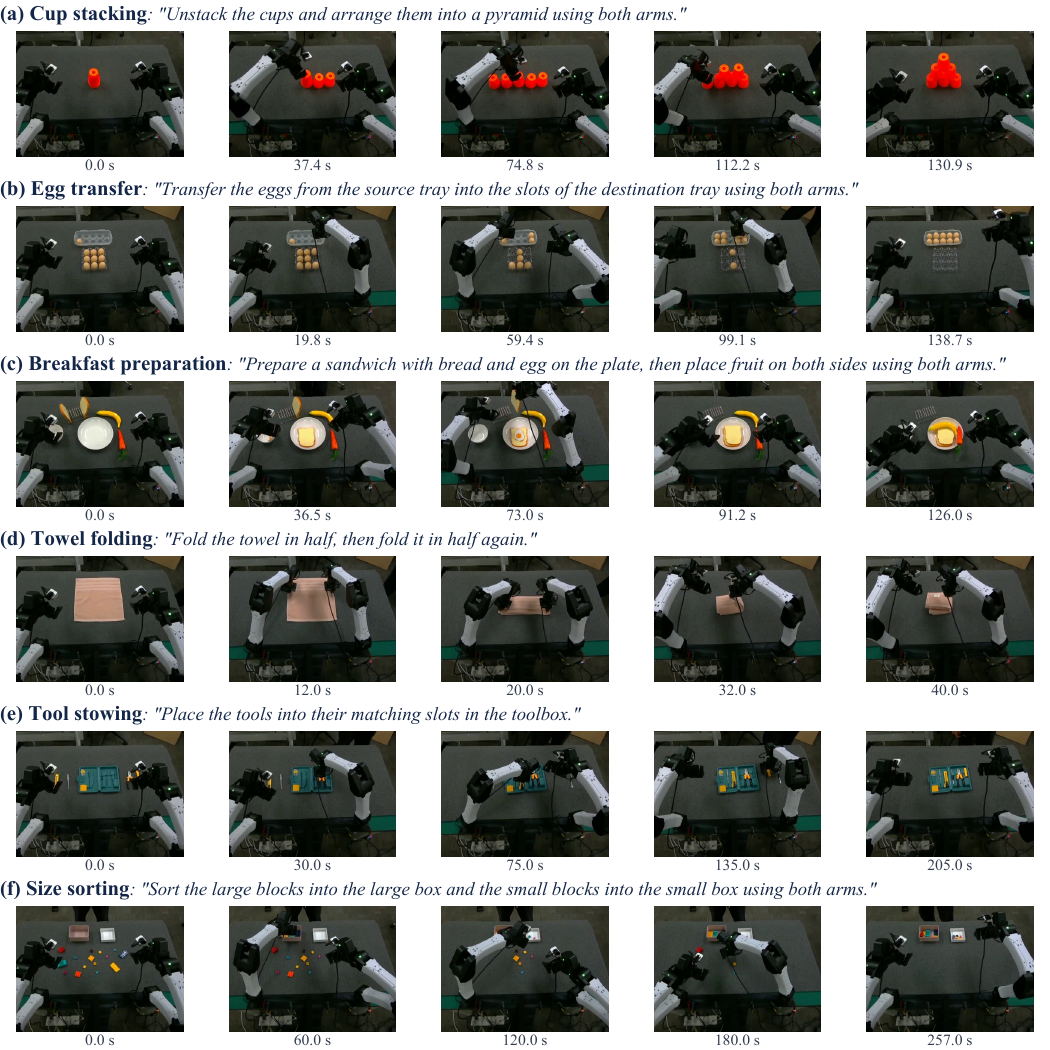}
    \caption{\textbf{Real-world rollout progressions.} Representative \method{} rollouts for cup stacking, egg transfer, breakfast preparation, towel folding, tool stowing, and size sorting, each labeled with its language instruction. Each row shows five chronological observations with elapsed wall-clock time from the start of execution.}
    \label{fig:real-world-progressions}
\end{figure*}

For the real-world comparison, $\pi_{0.5}$, ImageWAM, and \method{} are each jointly trained on six bimanual manipulation tasks: cup stacking, towel folding, egg transfer, breakfast preparation, tool stowing, and size sorting. Every task is evaluated over 15 trials. Cup stacking, towel folding, egg transfer, and breakfast preparation are scored by full-task completion, whereas tool stowing (4 tools per trial) and size sorting (15 blocks per trial) are scored per item. The overall score is the unweighted mean across all six tasks. Figure~\ref{fig:real-world-progressions} visualizes representative rollouts.

Table~\ref{tab:real-data} summarizes the six-task training dataset, recorded at 30\,Hz. Hours report total demonstration duration per task, and Sec./ep.\ reports the mean duration per episode.

\begin{table}[H]
\centering
\caption{\textbf{Real-world training dataset statistics.} All demonstrations were recorded at 30\,Hz across the six bimanual manipulation tasks.}
\label{tab:real-data}
\footnotesize
\setlength{\tabcolsep}{6pt}
\renewcommand{\arraystretch}{1.14}
\begin{tabularx}{0.66\textwidth}{Xrrr}
\toprule
Task & Episodes & Hours & Sec./ep. \\
\midrule
Cup stacking & 125 & 3.9 & 113.5 \\
Towel folding & 185 & 1.9 & 36.3 \\
Egg transfer & 132 & 4.3 & 118.0 \\
Breakfast preparation & 177 & 4.0 & 81.4 \\
Tool stowing & 154 & 4.9 & 114.4 \\
Size sorting & 150 & 3.9 & 94.2 \\
\midrule
\rowcolor{oursrow}
\textbf{Total / Mean} & \textbf{923} & \textbf{23.0} & \textbf{89.5} \\
\bottomrule
\end{tabularx}
\end{table}

\subsection{Task Success Criteria}
\paragraph{Cup stacking.}
The robot unstacks a nested pile of cups into a horizontal row and then stacks them into a pyramid. A trial succeeds if the entire sequence finishes within the time limit. Minor execution errors (e.g., missed grasps, tipped cups, or overlapping placements) are permitted if the policy autonomously recovers and completes the pyramid within the allotted time; otherwise, the trial fails.

\paragraph{Towel folding.}
The robot folds a flat towel in half, folds it in half a second time, and presses the folded cloth flat. A trial succeeds only if all three steps are completed cleanly; failure at any stage counts as a failed trial.

\paragraph{Egg transfer.}
The robot transfers eggs from a source carton into designated slots in a destination tray. A trial succeeds when all eggs are properly seated in the tray; rolling eggs or visibly misplaced placements count as failure.

\paragraph{Breakfast preparation.}
The robot sequentially places a slice of bread, a fried egg, a second slice of bread, a banana, and a carrot onto a plate. A trial succeeds if the full assembly completes within the time limit, including autonomous recovery from intermediate mistakes; any incomplete sequence counts as failure.

\paragraph{Tool stowing.}
Each trial presents four articulated tools to be placed into fitted organizer slots in a toolbox. We score the fraction of tools stowed successfully; if a tool fails, it is marked as a failure and an operator places it so the policy can proceed to the remaining tools. The reported score is the total number of stowed tools divided by 60 across 15 trials.

\paragraph{Size sorting.}
Each trial contains 15 blocks to be sorted by size into a large box and a small box. We report the mean fraction of correctly sorted blocks across 15 trials. A block counts as a failure if the robot fails to grasp it, selects the wrong box, or drops it outside the target box.

\section{Attention Analysis}
\label{sec:additional}
\label{sec:attention-diagnostics}
We apply an identical diagnostic procedure to ImageWAM and \method{} on randomized RoboTwin observations (Figure~\ref{fig:robotwin-attention}), holding the current RGB inputs, proprioceptive state, instruction, and sampling seed fixed within each pair to compare the two methods under matched inputs.

For Figure~\ref{fig:robotwin-attention}, we adapt attention rollout~\cite{abnar2020quantifying} with gradient-weighted attention propagation~\cite{chefer2021generic}, combining head-averaged positive attention--gradient products with additive residual transport across the image backbone, shared query updater, and Action Expert. Attribution targets the first action's left-gripper component at the final denoising step with earlier action latents held fixed. Routes are grouped by their entry edge into the action stream and normalized over all three camera views (only the head view is displayed).

\section{ImageWAM Evaluation on LIBERO-Plus}
\label{sec:imagewam-libero-plus}
The ImageWAM baseline in Table~\ref{tab:comparison-libero-plus} reflects our evaluation of the officially released FLUX.2 [klein] 4B checkpoint under the exact same protocol as \method{}, yielding an overall success rate of 81.6\% (used in Sec.~\ref{sec:ablations-robustness}) rather than the 83.1\% originally reported in~\cite{imagewam}.

As clarified by the ImageWAM authors in repository Issue~\#3,\footnote{\url{https://github.com/yuyangalin/ImageWAM/issues/3}} the originally reported 4B figure was evaluated on a 15\% subset using sampling seed 42, and the released evaluation script defaulted to a text context length of 128 whereas the 4B checkpoint was trained with 512. Evaluating the official checkpoint under our full matched evaluation setup ensures a fair comparison with \method{}.

\clearpage

\newsavebox{\robotwinfulltablebox}
\begin{table*}[p]
\centering
\tiny
\caption{\textbf{Per-task results on RoboTwin 2.0 Full.} Success rates (\%) are reported under clean and randomized evaluation conditions.}
\label{tab:robotwin-full-detail}
\setlength{\tabcolsep}{2.8pt}
\renewcommand{\arraystretch}{1.08}
\sbox{\robotwinfulltablebox}{%
\begin{tabular}{@{}lcccccccccccccc@{}}
\toprule
\multirow{2}{*}{Task} & \multicolumn{2}{c}{\makecell[c]{\textbf{\method{}}\\\textbf{(FLUX.2 4B)}}} & \multicolumn{2}{c}{\makecell[c]{\textbf{ImageWAM}\\\textbf{(FLUX.2 4B)}}} & \multicolumn{2}{c}{\textbf{Fast-WAM}} & \multicolumn{2}{c}{\makecell[c]{\textbf{FLEX-$\pi$}\\\textbf{(full joint)}}} & \multicolumn{2}{c}{\textbf{LingBot-VA}} & \multicolumn{2}{c}{\textbf{$\pi_{0.5}$}} & \multicolumn{2}{c}{\textbf{Motus}} \\
\cmidrule(lr){2-3}\cmidrule(lr){4-5}\cmidrule(lr){6-7}\cmidrule(lr){8-9}\cmidrule(lr){10-11}\cmidrule(lr){12-13}\cmidrule(lr){14-15}
 & Clean & Rand. & Clean & Rand. & Clean & Rand. & Clean & Rand. & Clean & Rand. & Clean & Rand. & Clean & Rand. \\
\midrule    
Adjust Bottle & \textbf{100} & \underline{99} & \textbf{100} & \underline{99} & \textbf{100} & \textbf{100} & \underline{98} & 97 & \underline{98} & 94 & \textbf{100} & \underline{99} & 89 & 93 \\
\rowcolor{zebrarow}
Beat Block Hammer & \textbf{99} & \textbf{99} & \underline{98} & \textbf{99} & \textbf{99} & 97 & \textbf{99} & 93 & 96 & \underline{98} & 96 & 93 & 95 & 88 \\
Blocks Ranking RGB & \textbf{100} & \underline{99} & 96 & \underline{99} & \textbf{100} & \textbf{100} & \textbf{100} & 93 & \underline{99} & 98 & 92 & 85 & \underline{99} & 97 \\
\rowcolor{zebrarow}
Blocks Ranking Size & \underline{95} & 97 & \textbf{96} & \textbf{100} & 94 & \underline{98} & 82 & 91 & 94 & 96 & 49 & 26 & 75 & 63 \\
Click Alarmclock & \textbf{100} & \textbf{100} & 98 & \textbf{100} & \textbf{100} & \textbf{100} & 91 & \underline{91} & \underline{99} & \textbf{100} & 98 & 89 & \textbf{100} & \textbf{100} \\
\rowcolor{zebrarow}
Click Bell & \textbf{100} & \textbf{100} & \textbf{100} & \textbf{100} & \textbf{100} & \textbf{100} & 94 & \underline{97} & \textbf{100} & \textbf{100} & \underline{99} & 66 & \textbf{100} & \textbf{100} \\
Dump Bin Bigbin & \underline{96} & 95 & \underline{96} & 90 & \textbf{97} & \underline{96} & 87 & 86 & 89 & \underline{96} & 92 & \textbf{97} & 95 & 91 \\
\rowcolor{zebrarow}
Grab Roller & \textbf{100} & \textbf{100} & \textbf{100} & \textbf{100} & \textbf{100} & \textbf{100} & \textbf{100} & \textbf{100} & \textbf{100} & \textbf{100} & \textbf{100} & \textbf{100} & \textbf{100} & \textbf{100} \\
Handover Block & \underline{97} & \textbf{96} & 96 & \underline{95} & 95 & 81 & \textbf{99} & \underline{95} & \textbf{99} & 78 & 66 & 57 & 86 & 73 \\
\rowcolor{zebrarow}
Handover Mic & \textbf{100} & \underline{99} & \textbf{100} & \textbf{100} & \underline{99} & \textbf{100} & \textbf{100} & \underline{99} & 94 & 96 & 98 & 97 & 78 & 63 \\
Hanging Mug & \textbf{92} & \textbf{89} & 74 & 84 & 58 & 62 & \underline{84} & \underline{86} & 40 & 28 & 18 & 17 & 38 & 38 \\
\rowcolor{zebrarow}
Lift Pot & \textbf{100} & \textbf{100} & \textbf{100} & \textbf{100} & \textbf{100} & \textbf{100} & \underline{99} & \textbf{100} & \textbf{100} & \underline{99} & 96 & 85 & 96 & \underline{99} \\
Move Can Pot & \underline{97} & 97 & 96 & \underline{98} & 90 & 88 & \textbf{98} & \textbf{100} & 94 & 97 & 51 & 55 & 34 & 74 \\
\rowcolor{zebrarow}
Move Pillbottle Pad & \underline{99} & \underline{99} & 98 & \textbf{100} & \textbf{100} & \underline{99} & \textbf{100} & \textbf{100} & \underline{99} & \underline{99} & 84 & 61 & 93 & 96 \\
Move Playingcard Away & \underline{98} & \textbf{100} & \textbf{100} & \underline{99} & \textbf{100} & \textbf{100} & \textbf{100} & \textbf{100} & \textbf{100} & \underline{99} & 96 & 84 & \textbf{100} & 96 \\
\rowcolor{zebrarow}
Move Stapler Pad & 84 & 84 & 67 & 60 & 77 & 64 & \textbf{96} & \textbf{94} & \underline{91} & 79 & 56 & 42 & 83 & \underline{85} \\
Open Laptop & \textbf{98} & \underline{98} & \textbf{98} & \underline{98} & \textbf{98} & \textbf{100} & \textbf{98} & \textbf{100} & 92 & 94 & 90 & 96 & \underline{95} & 91 \\
\rowcolor{zebrarow}
Open Microwave & 94 & \textbf{97} & \textbf{97} & \underline{94} & 62 & 45 & 70 & 77 & 82 & 86 & 34 & 77 & \underline{95} & 91 \\
Pick Diverse Bottles & 88 & 88 & 84 & 88 & 80 & 85 & \textbf{92} & \textbf{93} & 89 & 82 & 81 & 71 & \underline{90} & \underline{91} \\
\rowcolor{zebrarow}
Pick Dual Bottles & \underline{97} & 96 & 96 & 98 & \textbf{100} & 96 & \textbf{100} & \textbf{100} & \textbf{100} & \underline{99} & 93 & 63 & 96 & 90 \\
Place A2B Left & \textbf{99} & \textbf{100} & 95 & 93 & 95 & 93 & 95 & \underline{97} & \underline{97} & 93 & 87 & 82 & 88 & 79 \\
\rowcolor{zebrarow}
Place A2B Right & \textbf{97} & \textbf{99} & \underline{96} & 94 & 93 & \textbf{99} & \textbf{97} & \underline{97} & \textbf{97} & 95 & 87 & 84 & 91 & 87 \\
Place Bread Basket & 94 & \textbf{97} & \underline{96} & 92 & 91 & 93 & \underline{96} & 94 & \textbf{97} & \underline{95} & 77 & 64 & 91 & 94 \\
\rowcolor{zebrarow}
Place Bread Skillet & \underline{91} & \underline{94} & 90 & 89 & 90 & 93 & 90 & \textbf{95} & \textbf{95} & 90 & 85 & 66 & 86 & 83 \\
Place Burger Fries & \textbf{98} & \textbf{100} & 95 & \textbf{100} & 96 & \underline{99} & \textbf{98} & \underline{99} & \underline{97} & 95 & 94 & 87 & \textbf{98} & 98 \\
\rowcolor{zebrarow}
Place Can Basket & 70 & 68 & 74 & 72 & 71 & 69 & \underline{80} & \textbf{84} & \textbf{81} & \textbf{84} & 62 & 62 & \textbf{81} & \underline{76} \\
Place Cans Plasticbox & 98 & 98 & \underline{99} & 97 & \underline{99} & 96 & \textbf{100} & \textbf{100} & \textbf{100} & \underline{99} & 94 & 84 & 98 & 94 \\
\rowcolor{zebrarow}
Place Container Plate & \textbf{99} & \underline{99} & \underline{98} & 98 & 96 & \textbf{100} & 97 & \textbf{100} & \textbf{99} & 97 & \textbf{99} & 95 & \underline{98} & \underline{99} \\
Place Dual Shoes & 84 & 88 & 81 & 81 & \underline{94} & 88 & \textbf{98} & \textbf{93} & \underline{94} & \underline{89} & 75 & 75 & 93 & 87 \\
\rowcolor{zebrarow}
Place Empty Cup & \textbf{100} & \textbf{100} & \textbf{100} & \textbf{100} & \textbf{100} & \textbf{100} & \textbf{100} & \textbf{100} & \textbf{100} & \textbf{100} & \textbf{100} & \underline{99} & \underline{99} & 98 \\
Place Fan & \underline{97} & 92 & 95 & 94 & 96 & \underline{96} & \underline{97} & \textbf{97} & \textbf{99} & 93 & 87 & 85 & 91 & 87 \\
\rowcolor{zebrarow}
Place Mouse Pad & \textbf{98} & 94 & 84 & 93 & 83 & 89 & \textbf{98} & \textbf{98} & \underline{93} & \underline{96} & 60 & 39 & 66 & 68 \\
Place Object Basket & 86 & 85 & 86 & 83 & \underline{89} & \underline{88} & 84 & \textbf{92} & \textbf{91} & \underline{88} & 80 & 76 & 81 & 87 \\
\rowcolor{zebrarow}
Place Object Scale & \underline{96} & \textbf{98} & \textbf{97} & 96 & 90 & \underline{97} & \underline{96} & \textbf{98} & \underline{96} & 95 & 86 & 80 & 88 & 85 \\
Place Object Stand & 96 & 97 & \underline{98} & \underline{98} & 90 & 94 & 97 & \textbf{99} & \textbf{99} & 96 & 91 & 85 & \underline{98} & 97 \\
\rowcolor{zebrarow}
Place Phone Stand & \textbf{100} & \underline{99} & \textbf{100} & \textbf{100} & 97 & \underline{99} & \underline{98} & \textbf{100} & 97 & 97 & 81 & 81 & 87 & 86 \\
Place Shoe & 94 & 97 & 97 & 95 & 96 & \textbf{99} & \underline{98} & \underline{98} & \underline{98} & \underline{98} & 92 & 93 & \textbf{99} & 97 \\
\rowcolor{zebrarow}
Press Stapler & \textbf{97} & \textbf{100} & \textbf{97} & \textbf{100} & 90 & 97 & \textbf{97} & \underline{98} & 85 & 82 & 87 & 83 & \underline{93} & \underline{98} \\
Put Bottles Dustbin & \underline{96} & \underline{95} & \textbf{97} & 91 & 95 & 90 & 95 & \textbf{96} & 87 & 91 & 84 & 79 & 81 & 79 \\
\rowcolor{zebrarow}
Put Object Cabinet & \textbf{94} & \underline{93} & \underline{91} & 89 & \textbf{94} & 89 & 90 & \textbf{95} & 85 & 87 & 80 & 79 & 88 & 71 \\
Rotate QRcode & 90 & 90 & 87 & \textbf{92} & \underline{93} & 89 & 84 & 88 & \textbf{96} & \underline{91} & 89 & 87 & 89 & 73 \\
\rowcolor{zebrarow}
Scan Object & 92 & \textbf{93} & \underline{94} & 90 & 89 & \underline{92} & 92 & 87 & \textbf{96} & 91 & 72 & 65 & 67 & 66 \\
Shake Bottle & \underline{99} & \underline{99} & \textbf{100} & \textbf{100} & \textbf{100} & \textbf{100} & \textbf{100} & \underline{99} & \textbf{100} & 97 & \underline{99} & 97 & \textbf{100} & 97 \\
\rowcolor{zebrarow}
Shake Bottle Horizontally & \textbf{100} & \textbf{100} & \textbf{100} & \textbf{100} & \textbf{100} & \textbf{100} & \textbf{100} & \underline{99} & \textbf{100} & \underline{99} & \underline{99} & \underline{99} & \textbf{100} & 98 \\
Stack Blocks Three & \underline{98} & \textbf{98} & 96 & \underline{97} & 95 & \underline{97} & \textbf{99} & 96 & \textbf{99} & \textbf{98} & 91 & 76 & 91 & 95 \\
\rowcolor{zebrarow}
Stack Blocks Two & \textbf{100} & \underline{99} & \underline{99} & \textbf{100} & \textbf{100} & \textbf{100} & \textbf{100} & \textbf{100} & \textbf{100} & 98 & 97 & \textbf{100} & \textbf{100} & 98 \\
Stack Bowls Three & \textbf{86} & 82 & 78 & 83 & \underline{80} & 81 & 77 & \underline{84} & \textbf{86} & 83 & 77 & 71 & 79 & \textbf{87} \\
\rowcolor{zebrarow}
Stack Bowls Two & \underline{95} & \underline{97} & 94 & \underline{97} & 92 & \textbf{98} & \underline{95} & \textbf{98} & 94 & \textbf{98} & \underline{95} & 96 & \textbf{98} & \textbf{98} \\
Stamp Seal & 90 & 91 & 79 & 84 & 90 & 94 & \textbf{99} & \textbf{99} & \underline{96} & \underline{97} & 79 & 55 & 93 & 92 \\
\rowcolor{zebrarow}
Turn Switch & 77 & \textbf{82} & 77 & \underline{79} & 61 & 59 & \underline{81} & 77 & 44 & 45 & 62 & 54 & \textbf{84} & 78 \\
\midrule
\rowcolor{oursrow}
\textbf{Average} & \textbf{94.90} & \textbf{95.12} & 93.20 & 93.56 & 91.88 & 91.78 & \underline{94.30} & \underline{94.80} & 92.90 & 91.50 & 82.74 & 76.76 & 88.66 & 87.02 \\
\bottomrule
\end{tabular}
}
\makeatletter
\Gscale@div\robotwintablescale{\textwidth}{\wd\robotwinfulltablebox}
\global\let\robotwintablescale\robotwintablescale
\makeatother
\scalebox{\robotwintablescale}{\usebox{\robotwinfulltablebox}}
\end{table*}
\clearpage

\begin{table*}[p]
\centering
\tiny
\caption{\textbf{Per-task results on RoboTwin 2.0 Clean-to-Random.} Success rates (\%) are reported under clean and randomized evaluation conditions after training on clean demonstrations.}
\label{tab:robotwin-c2r-detail}
\setlength{\tabcolsep}{2.8pt}
\renewcommand{\arraystretch}{1.08}
\scalebox{\robotwintablescale}{%
\begin{tabular}{@{}lcccccccccccc@{}}
\toprule
\multirow{2}{*}{Task} & \multicolumn{2}{c}{\textbf{\method{}}} & \multicolumn{2}{c}{\textbf{ImageWAM}} & \multicolumn{2}{c}{\textbf{$\pi_{0.5}$}} & \multicolumn{2}{c}{\textbf{X-WAM}} & \multicolumn{2}{c}{\textbf{4D-WAM}} & \multicolumn{2}{c}{\textbf{Fast-WAM}} \\
\cmidrule(lr){2-3}\cmidrule(lr){4-5}\cmidrule(lr){6-7}\cmidrule(lr){8-9}\cmidrule(lr){10-11}\cmidrule(lr){12-13}
 & Clean & Rand. & Clean & Rand. & Clean & Rand. & Clean & Rand. & Clean & Rand. & Clean & Rand. \\
\midrule
Adjust Bottle & \textbf{100} & \underline{75} & 98 & 50 & \underline{99} & \textbf{77} & 92 & 13 & 80 & 36 & \underline{99} & 0 \\
\rowcolor{zebrarow}
Beat Block Hammer & \textbf{94} & \underline{60} & \underline{93} & \textbf{64} & \underline{93} & 23 & 92 & 21 & 92 & 28 & 82 & 0 \\
Blocks Ranking RGB & \underline{96} & \textbf{98} & \textbf{97} & \underline{77} & 72 & 44 & 85 & 36 & 88 & 10 & 81 & 0 \\
\rowcolor{zebrarow}
Blocks Ranking Size & \underline{67} & \textbf{59} & \textbf{71} & \underline{57} & 45 & 21 & 33 & 0 & 54 & 8 & 52 & 0 \\
Click Alarmclock & \textbf{100} & \underline{83} & \textbf{100} & \textbf{88} & 65 & 50 & 95 & 61 & \textbf{100} & 70 & \underline{99} & 60 \\
\rowcolor{zebrarow}
Click Bell & \textbf{100} & 61 & \textbf{100} & \textbf{84} & 31 & 35 & \textbf{100} & 52 & \underline{98} & \underline{77} & 95 & 7 \\
Dump Bin Bigbin & \textbf{98} & 83 & \underline{96} & \textbf{86} & 92 & \underline{84} & 88 & 40 & 86 & 57 & 94 & 0 \\
\rowcolor{zebrarow}
Grab Roller & \textbf{100} & \underline{78} & \underline{99} & 66 & \underline{99} & \textbf{97} & 95 & 16 & \textbf{100} & 68 & \textbf{100} & 2 \\
Handover Block & \textbf{93} & \textbf{64} & 63 & \underline{44} & 30 & 12 & 21 & 21 & \underline{79} & 25 & 66 & 0 \\
\rowcolor{zebrarow}
Handover Mic & \textbf{98} & \textbf{46} & 69 & \underline{40} & \textbf{98} & 6 & \underline{97} & 8 & 85 & 13 & \textbf{98} & 0 \\
Hanging Mug & \underline{53} & \textbf{27} & 21 & \underline{24} & 17 & 14 & 17 & 0 & \textbf{70} & 10 & 14 & 0 \\
\rowcolor{zebrarow}
Lift Pot & \textbf{100} & \textbf{68} & \underline{99} & 30 & 98 & \underline{35} & 77 & 0 & 88 & 9 & 91 & 0 \\
Move Can Pot & 88 & \textbf{47} & 85 & 27 & 60 & 10 & 77 & \underline{34} & \underline{94} & 17 & \textbf{96} & 0 \\
\rowcolor{zebrarow}
Move Pillbottle Pad & \textbf{98} & \textbf{59} & \underline{97} & \textbf{59} & 46 & \underline{32} & 78 & 26 & 89 & 22 & 92 & 0 \\
Move Playingcard Away & \textbf{98} & \textbf{86} & \underline{97} & 60 & 84 & \underline{65} & 83 & 11 & 82 & 35 & 94 & 0 \\
\rowcolor{zebrarow}
Move Stapler Pad & \textbf{54} & \textbf{21} & 25 & 14 & 20 & \underline{18} & 26 & 2 & \underline{43} & 12 & 33 & 0 \\
Open Laptop & \textbf{94} & \underline{72} & 89 & 69 & \underline{93} & 68 & 72 & 6 & 85 & \textbf{82} & 84 & 0 \\
\rowcolor{zebrarow}
Open Microwave & 67 & 36 & \underline{71} & \underline{42} & \textbf{86} & \textbf{59} & 9 & 24 & 68 & 12 & 17 & 0 \\
Pick Diverse Bottles & \textbf{83} & \textbf{65} & 75 & \underline{55} & 66 & 42 & 74 & 26 & \underline{77} & 44 & 72 & 0 \\
\rowcolor{zebrarow}
Pick Dual Bottles & \textbf{100} & \textbf{75} & 98 & \underline{71} & 72 & 57 & \underline{99} & 55 & 92 & 67 & 87 & 1 \\
Place A2B Left & \textbf{92} & \textbf{64} & 69 & 40 & \underline{71} & \underline{49} & 70 & 18 & \underline{71} & 19 & 69 & 0 \\
\rowcolor{zebrarow}
Place A2B Right & \textbf{89} & \textbf{57} & \underline{74} & 38 & 68 & 39 & 59 & 21 & 62 & \underline{42} & 60 & 0 \\
Place Bread Basket & \textbf{91} & \textbf{88} & \underline{90} & \underline{85} & 63 & 60 & 82 & 40 & 87 & 77 & 87 & 0 \\
\rowcolor{zebrarow}
Place Bread Skillet & \textbf{96} & \underline{73} & \underline{91} & \textbf{79} & 76 & 47 & 82 & 37 & 82 & 69 & 85 & 1 \\
Place Burger Fries & \underline{97} & \textbf{97} & \textbf{100} & \textbf{97} & 82 & \underline{86} & 93 & 36 & 86 & 72 & \underline{97} & 0 \\
\rowcolor{zebrarow}
Place Can Basket & 73 & \textbf{56} & 75 & \underline{55} & 56 & 10 & 63 & 15 & \textbf{83} & 51 & \underline{77} & 0 \\
Place Cans Plasticbox & \textbf{99} & \textbf{94} & \textbf{99} & \underline{92} & 31 & 73 & 93 & 36 & 94 & 48 & \underline{98} & 0 \\
\rowcolor{zebrarow}
Place Container Plate & \underline{99} & \textbf{89} & \underline{99} & \underline{85} & 92 & 72 & 95 & 52 & 93 & 82 & \textbf{100} & 0 \\
Place Dual Shoes & \textbf{70} & \textbf{63} & 66 & 24 & \textbf{70} & \underline{45} & 38 & 7 & 66 & 19 & \underline{67} & 0 \\
\rowcolor{zebrarow}
Place Empty Cup & \underline{99} & \textbf{87} & \textbf{100} & 61 & 96 & \underline{80} & 93 & 36 & \underline{99} & 70 & 90 & 0 \\
Place Fan & \textbf{85} & \textbf{57} & \underline{67} & \underline{37} & 66 & 26 & 47 & 6 & 48 & 17 & 59 & 0 \\
\rowcolor{zebrarow}
Place Mouse Pad & \underline{57} & \textbf{37} & \textbf{62} & \underline{34} & 34 & 31 & 49 & 6 & 54 & 9 & 55 & 0 \\
Place Object Basket & \textbf{85} & \textbf{77} & 56 & \underline{73} & 70 & 29 & 52 & 20 & 75 & 40 & \underline{82} & 0 \\
\rowcolor{zebrarow}
Place Object Scale & \textbf{95} & \textbf{62} & \underline{84} & \underline{54} & 78 & 38 & 66 & 14 & 79 & 24 & 83 & 0 \\
Place Object Stand & \textbf{96} & \textbf{88} & 90 & \underline{81} & 85 & 47 & 83 & 28 & 87 & 42 & \underline{94} & 0 \\
\rowcolor{zebrarow}
Place Phone Stand & \textbf{94} & \textbf{57} & 79 & \underline{44} & 65 & 37 & 35 & 4 & 84 & 21 & \underline{85} & 0 \\
Place Shoe & \textbf{93} & \textbf{75} & 82 & \underline{53} & 82 & 50 & 76 & 31 & 83 & 52 & \underline{87} & 0 \\
\rowcolor{zebrarow}
Press Stapler & 88 & 53 & \textbf{92} & \textbf{75} & 79 & \underline{63} & 85 & 20 & 76 & 61 & \underline{90} & 0 \\
Put Bottles Dustbin & \textbf{90} & \textbf{85} & \underline{84} & \underline{73} & 78 & 50 & 71 & 21 & 68 & 19 & 55 & 0 \\
\rowcolor{zebrarow}
Put Object Cabinet & 55 & \underline{30} & \underline{56} & 29 & 39 & 22 & 28 & 16 & \textbf{69} & \textbf{47} & 42 & 0 \\
Rotate QRcode & \textbf{90} & 18 & 79 & 5 & \textbf{90} & \underline{21} & 47 & 2 & 79 & \textbf{67} & \underline{81} & 3 \\
\rowcolor{zebrarow}
Scan Object & \textbf{76} & \underline{46} & 71 & 42 & 49 & 40 & 32 & 14 & \underline{73} & \textbf{53} & 68 & 0 \\
Shake Bottle & \underline{99} & \underline{96} & \textbf{100} & 94 & \textbf{100} & \textbf{100} & 96 & 67 & 98 & 72 & \textbf{100} & 10 \\
\rowcolor{zebrarow}
Shake Bottle Horizontally & \textbf{100} & \underline{97} & \textbf{100} & 96 & \underline{99} & \textbf{100} & 97 & 66 & 98 & 70 & \textbf{100} & 9 \\
Stack Blocks Three & \underline{97} & \textbf{94} & \textbf{100} & 74 & 77 & 25 & 72 & 26 & 95 & \underline{82} & 89 & 0 \\
\rowcolor{zebrarow}
Stack Blocks Two & \textbf{100} & \textbf{99} & \textbf{100} & \underline{97} & 91 & 43 & 97 & 76 & \textbf{100} & 34 & \underline{99} & 0 \\
Stack Bowls Three & \textbf{84} & \textbf{75} & 66 & \underline{63} & 81 & 52 & 69 & 38 & \underline{82} & 27 & 80 & 0 \\
\rowcolor{zebrarow}
Stack Bowls Two & \textbf{96} & \textbf{90} & \underline{94} & \underline{86} & \underline{94} & 70 & 93 & 67 & \underline{94} & 81 & \underline{94} & 0 \\
Stamp Seal & 62 & \textbf{32} & \underline{71} & \underline{28} & 53 & 21 & 58 & 16 & \textbf{74} & 10 & 30 & 0 \\
\rowcolor{zebrarow}
Turn Switch & 64 & \textbf{54} & 55 & \underline{40} & 52 & 25 & \underline{68} & 4 & \textbf{86} & 11 & 39 & 3 \\
\midrule
\rowcolor{oursrow}
\textbf{Average} & \textbf{87.84} & \textbf{67.06} & \underline{81.88} & \underline{58.82} & 70.66 & 46.00 & 69.98 & 25.84 & 81.50 & 41.80 & 77.76 & 1.92 \\
\bottomrule
\end{tabular}
}
\end{table*}
\clearpage

\clearpage

\end{document}